\documentclass[10pt,twocolumn,letterpaper]{article}

\usepackage[pagenumbers]{cvpr} 

\usepackage{graphicx}
\usepackage{amsmath}
\usepackage{amssymb}
\usepackage{booktabs}

\usepackage{enumitem}
\usepackage{gensymb}
\usepackage{multirow}

\usepackage[pagebackref,breaklinks,colorlinks]{hyperref}

\usepackage[capitalize]{cleveref}
\crefname{section}{Sec.}{Secs.}
\Crefname{section}{Section}{Sections}
\Crefname{table}{Table}{Tables}
\crefname{table}{Tab.}{Tabs.}

\DeclareMathOperator*{\argmin}{arg\,min}

\newcommand{\SL}{\mathrm{SL}}
\newcommand{\SE}{\mathrm{SE}}
\newcommand{\SO}{\mathrm{SO}}
\newcommand{\lsl}{\mathfrak{sl}}
\newcommand{\se}{\mathfrak{se}}
\newcommand{\so}{\mathfrak{so}}
\newcommand{\bA}{\boldsymbol{A}}
\newcommand{\bH}{\boldsymbol{H}}
\newcommand{\bU}{\boldsymbol{U}}
\newcommand{\bS}{\boldsymbol{S}}
\newcommand{\bV}{\boldsymbol{V}}
\newcommand{\wu}{{\x^j}^{\prime}_{\hspace{-2.3pt}1}}
\newcommand{\wv}{{\x^j}^{\prime}_{\hspace{-2.3pt}2}}
\newcommand{\ou}{{\x^j}_{\hspace{-2.3pt}1}}
\newcommand{\ov}{{\x^j}_{\hspace{-2.3pt}2}}

\newcommand{\I}{\mathcal{I}}

\newcommand{\x}{\mathbf{x}}
\newcommand{\0}{\mathbf{0}}
\newcommand{\W}{\mathcal{W}}

\newcommand{\bTheta}{\boldsymbol{\Theta}}
\renewcommand{\u}{\mathbf{u}}

\renewcommand{\c}{\mathbf{c}}
\newcommand{\y}{\mathbf{y}}

\newcommand{\pderiv}[2]{\frac{\partial #1}{\partial #2}}

\newcommand{\SFM}{S\textit{f}M\xspace}

\newcommand{\eye}{\mathbf{I}}
\newcommand{\Real}{\mathbb{R}}

\newcommand{\bPhi}{\boldsymbol{\Phi}}

\newcommand{\bell}{\boldsymbol{\ell}}

\newcommand{\fr}{f_{\hspace{-1pt}{\R}}}
\newcommand{\fw}{f_{\hspace{-1pt}{\W}}}

\newcommand{\T}{\mathbf{T}}
\newcommand{\bt}{\mathbf{t}}
\newcommand{\bR}{\mathbf{R}}
\newcommand{\R}{\mathcal{R}}
\def\wrt{w.r.t\onedot}

\def\confName{CVPR}
\def\confYear{2023}

\begin{document}

\title{Local-to-Global Registration for Bundle-Adjusting Neural Radiance Fields}

\author{Yue Chen$^{1}$\footnotemark[1] \qquad Xingyu Chen$^{1}$\footnotemark[1] \qquad Xuan Wang$^{2}$\footnotemark[2] \qquad  Qi Zhang$^3$ \\
\qquad Yu Guo$^1$\footnotemark[2] \qquad Ying Shan$^3$ \qquad Fei Wang$^1$ \vspace{3pt}\\
$^{1}$Xi'an Jiaotong University\ \ \ \  $^{2}$Ant Group\ \ \ \  $^{3}$Tencent AI Lab\\
}

\maketitle

\begin{abstract}
Neural Radiance Fields (NeRF) have achieved photorealistic novel views synthesis; however, the requirement of accurate camera poses limits its application.
Despite analysis-by-synthesis extensions for jointly learning neural 3D representations and registering camera frames exist, 
they are susceptible to suboptimal solutions if poorly initialized.
We propose \emph{L2G-NeRF}, a Local-to-Global registration method for bundle-adjusting Neural Radiance Fields:
first, a pixel-wise flexible alignment, followed by a frame-wise constrained parametric alignment.
Pixel-wise local alignment is learned in an unsupervised way via a deep network which optimizes photometric reconstruction errors.
Frame-wise global alignment is performed using differentiable parameter estimation solvers on the pixel-wise correspondences
to find a global transformation.
Experiments on synthetic and real-world data show that our method outperforms the current state-of-the-art in terms of high-fidelity reconstruction and resolving large camera pose misalignment.
Our module is an easy-to-use plugin that can be applied to NeRF variants and other neural field applications.
The Code and supplementary materials are available at \href{https://rover-xingyu.github.io/L2G-NeRF/}{https://rover-xingyu.github.io/L2G-NeRF/}.

\renewcommand{\thefootnote}{\fnsymbol{footnote}} 
\footnotetext[1]{Authors contributed equally to this work.}
\footnotetext[2]{Corresponding Author.}

\end{abstract}

\section{Introduction}
\label{sec:intro}
Recent success with neural fields~\cite{xie2022neural} has caused a resurgence of interest in visual computing problems, where coordinate-based neural networks that represent a field gain traction as a useful parameterization of 2D images~\cite{sitzmann2020implicit,bemana2020x,chen2021learning}, and 3D scenes~\cite{park2019deepsdf,mescheder2019occupancy,mildenhall2020nerf}. Commonly, these coordinates are warped to a global coordinate system by camera parameters obtained via computing homography, structure from motion (\SFM), or simultaneous localization and mapping (SLAM)~\cite{hartley2004} with off-the-shelf tools like COLMAP~\cite{schonberger2016structure}, before being fed to the neural fields.

\begin{figure}[t!]
    \centering
    \includegraphics[width=1\linewidth,page=1]{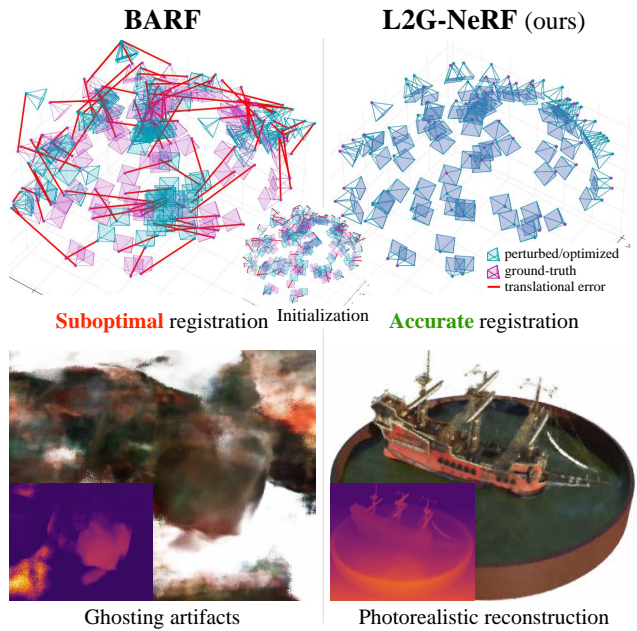}
    \caption{
        We present {\bf L2G-NeRF}, a new bundle-adjusting neural radiance fields --- employing local-to-global registration --- that is much more robust than the current state-of-the-art BARF~\cite{lin2021barf}.
    }
    \label{fig:teaser}
\end{figure}

This paper considers the generic problem of simultaneously 
\textbf{reconstructing} the neural fields from RGB images and
\textbf{registering} the given camera frames, which is known as a long-standing chicken-and-egg problem --- registration is needed to reconstruct the fields, and reconstruction is needed to register the cameras.

One straightforward way to solve this problem is to jointly optimize the camera parameters with the neural fields via backpropagation. Recent work can be broadly placed into two camps: parametric and non-parametric. Parametric methods~\cite{lin2021barf,wang2021nerf,jeong2021self,chng2022garf} directly optimize global geometric transformations (\eg rigid, homography). Non-parametric methods~\cite{kasten2021layered,nam2022neural} do not make any assumptions on the type of transformation, and attempt to directly optimize some pixel agreement metric (\eg brightness constancy constraint in optical flow and stereo). 

However, both approaches have flaws: parametric methods fail to minimize the photometric errors (falling into the suboptimal solutions) if poorly initialized, as shown in Fig.~\ref{fig:teaser}, while non-parametric methods have trouble dealing with large displacements (\eg although the photometric errors are minimized, the alignments do not obey the geometric constraint). It is natural, therefore, to consider a hybrid approach, combining the benefits of parametric and non-parametric methods together.

\begin{figure*}[t]
\centering
  \includegraphics[width=1\linewidth]{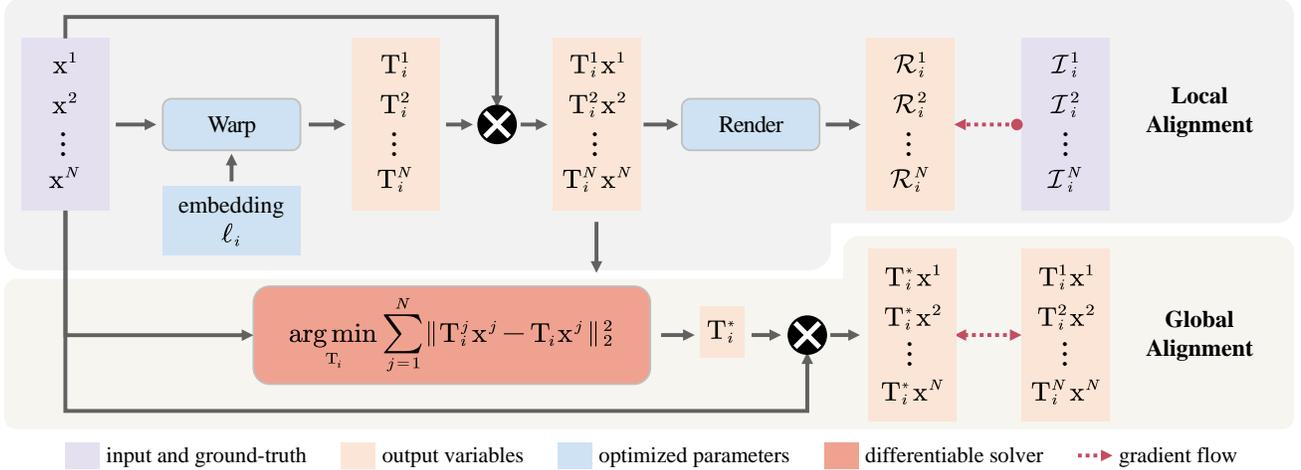}
   \caption{Overall pipeline of proposed framework. Our model has two main branches: 1) Based on query coordinates $\{\x^j\}_{j=1}^{N}$ and frame-dependent embeddings $\{\bell_i\}_{i=1}^{M}$, a warp neural field $\W$ constructs pixel-wise local transformations $\{\T^j_i\}_{i=1,j=1}^{M,N}$ and transforms query coordinates into a global coordinate system. Then the color can be rendered via a neural field $\R$ to minimize the photometric error between renderings $\{\R_i\}_{i=1}^{M}$ and images $\{\I_i\}_{i=1}^{M}$. 
   2) A differentiable parameter estimation solver produces frame-wise global transformations $\{\T^\ast_i\}_{i=1}^{M}$ condition on the pixel-wise correspondences. The query coordinates are then transformed to apply a global geometric constraint.}
\label{fig:pipe}
\end{figure*}

In this paper, we propose L2G-NeRF, a local-to-global process integrating parametric and non-parametric methods for bundle-adjusting neural radiance fields --- the joint problem of \emph{reconstructing} the neural fields and \emph{registering} the camera parameters, which can be regarded as a type of classic photometric bundle adjustment (BA)~\cite{delaunoy2014photometric,alismail2016photometric,lin2019photometric}. Fig.~\ref{fig:pipe} shows an overview. In the first non-parametric stage, we initialize the alignment by predicting a local transformation field for each pixel of the camera frames. This is achieved by self-supervised training of a deep network to optimize standard photometric reconstruction errors. In the second stage, differentiable parameter estimation solvers are applied to a set of pixel-wise correspondences to obtain a global alignment, which is then used to apply a soft constraint to the local alignment.
In summary, we present the following contributions:

\vspace{-2pt}
\begin{itemize}[leftmargin=24pt]
	\setlength\itemsep{0pt}
	\item We show that the optimization of bundle-adjusting neural fields is sensitive to initialization, and we present a simple yet effective strategy for local-to-global registration on neural fields.
	\item We introduce two differentiable parameter estimation solvers for rigid and homography transformation respectively, which play a crucial role in calculating the gradient flow from the global alignment to the local alignment.
	\item Our method is agnostic to the particular type of neural fields, specifically, we show that the local-to-global process works quite well in 2D neural images and 3D Neural Radiance Fields (NeRF)~\cite{mildenhall2020nerf}, allowing for applications such as image reconstruction and novel view synthesis.
\end{itemize}

\section{Related Work}

\noindent\textbf{\SFM and SLAM.}
\SFM~\cite{pollefeys1999self,pollefeys2004visual,snavely2006photo,snavely2008modeling,agarwal2011building} and SLAM~\cite{newcombe2011dtam,engel2014lsd,mur2015orb,yang2021asynchronous} systems attempt to simultaneously recover the 3D structure and the sensor poses from a set of input images. They reconstruct an explicit geometry (\eg point clouds) and estimate camera poses through image registration via associating feature correspondences~\cite{davison2007monoslam,mur2015orb} or minimizing photometric errors~\cite{alismail2016photometric,engel2017direct}, followed by BA~\cite{delaunoy2014photometric,alismail2016photometric,lin2019photometric}.

However, the explicit point clouds assume a diffuse surface, hence cannot model view-dependent appearance. And the sparse nature of point clouds also limits downstream vision tasks, such as photorealistic rendering. In contrast, L2G-NeRF encodes the scenes as coordinate-based neural fields, which is qualified for solving the high-fidelity visual computing problems.

\noindent\textbf{Neural Fields.} Recent advances in neural fields~\cite{xie2022neural}, which employ coordinate-based neural networks to parameterize physical properties of scenes or objects across space and time, have led to increased interest in solving visual computing problems, 
causing more accurate, higher fidelity, more expressive, and memory-efficient solutions. They have seen widespread success in problems such as image synthesis~\cite{sitzmann2020implicit,bemana2020x,chen2021learning}, 3D shape~\cite{park2019deepsdf,mescheder2019occupancy,chen2019learning}, view-dependent appearance~\cite{mildenhall2020nerf,chen2022hallucinated,huang2022hdr,niemeyer2021giraffe}, and animation of humans~\cite{peng2021neural,weng2022humannerf,chen2022uv}.

While these neural fields have achieved impressive results, the requirement of \emph{camera parameters} limits its application. 
We are able to get around the requirement with our proposed L2G-NeRF.

\noindent\textbf{Bundle-Adjusting Neural Fields.} Since neural fields are end-to-end differentiable, camera parameters can be jointly estimated with the neural fields. The optimization problem is known to be non-convex, and is reflected by NeRF-{}-~\cite{wang2021nerf}, in which the authors jointly optimize the scene and cameras for forward-facing scenes. 
Adversarial objective is utilized~\cite{meng2021gnerf} to relax forward-facing assumption and supports inward-facing $360\degree$ scenes. SCNeRF~\cite{jeong2021self} is further developed to learn the camera intrinsics. BARF~\cite{lin2021barf} shows that bundle-adjusting neural fields could benefit from coarse-to-fine registration. Recent approaches employ Gaussian activations~\cite{chng2022garf} or Sinusoidal activations~\cite{xia2022sinerf} to overcome local minima in optimization. 

Nevertheless, these parametric methods directly optimize global geometric transformations, which are prone to falling into suboptimal solutions if poorly initialized. Non-parametric methods~\cite{kasten2021layered,nam2022neural} directly optimize decent local transformations based on brightness constancy constraints, whereas they can not handle large displacements. We show that by combining the parametric and non-parametric methods together with a simple local-to-global process, we can achieve surprising anti-noise ability, allowing utilities for various NeRF extensions and other neural field applications.
\section{Approach} \label{sec:approach}

We first present the formulation of recovering the neural field jointly with camera parameters. Given a collection of images $\{\I_i\}_{i=1}^{M}$, we aim to jointly find the parameters $\bTheta$ of the neural field $\R$ and the camera parameters (geometric transformation matrices) $\{\T_i\}_{i=1}^{M}$ that minimize the photometric error between renderings and images. Let $\{\x^j\}_{j=1}^{N}$ be the query coordinates and $\I$ be the imaging function,
we formulate the problem as:
\begin{align} \label{eq:photometric_loss}
    \min_{\{\T_i\}_{i=1}^{M}, \bTheta} \;  \sum_{i=1}^{M} \sum^{N}_{j=1} \big( \big\| \R(\T_i \x^j; \bTheta) - \I_i(\x^j) \big\|_2^2\big) \;.
\end{align}

Gradient-based optimization is the preferred strategy to solve this nonlinear problem. 
Nevertheless, gradient-based registration is prone to finding suboptimal poses.
Therefore, we propose a simple yet effective strategy for local-to-global registration.
The key idea is to apply a pixel-wise flexible alignment that optimizes photometric reconstruction errors individually, followed by a frame-wise alignment to globally constrain the local geometric transformations, which acts like a soft extension of Eq.~\eqref{eq:photometric_loss}:
\begin{align} \label{eq:soft_photometric_loss}
    \min_{\{\T^j_i\}_{i=1,j=1}^{M,N}, \bTheta} \;  \sum_{i=1}^{M} \sum^{N}_{j=1} \big( & \big\|
    \R(\T^j_i \x^j; \bTheta) - \I_i(\x^j)\big\|_2^2
    \nonumber \\[-6pt]
    + \lambda & \big\|\T^j_i \x^j-\T^\ast_i \x^j\big\|_2^2\big) \;,
\end{align}
where the pixel-wise local transformations $\{\T^j_i\}_{i=1,j=1}^{M,N}$ are modeled by a warp neural field $\W$ parametrized by $\bPhi$, along with frame-dependent embeddings $\{\bell_i\}_{i=1}^{M}$:
\begin{align} \label{eq:local_transformations}
    \T^j_i = \W(\x^j;\bell_i,\bPhi) \;,
\end{align}
and the frame-wise global transformations $\{\T^\ast_i\}_{i=1}^{M}$ are solved by using differentiable parameter estimation solvers on the pixel-wise correspondences:
\begin{align}\label{eq:argmin}
\T^\ast_i = \argmin_{\T_i} & \sum^{N}_{j=1} \big\| \T^j_i \x^j-\T_i \x^j \big\|_2^2.
\end{align}
\subsection{Neural Image Alignment (2D)} \label{sec:image}
To develop intuition, we first consider the case of a 2D neural image alignment problem. More specifically, let  $\x \in \Real^2$ be the 2D pixel coordinates and $\I: \Real^2 \to \Real^3$, we aim to optimize a 2D neural field parameterized as the weights $\bTheta$ of a multilayer perceptron (MLP) $\fr: \Real^2 \to \Real^3$:
\begin{align} \label{eq:R2D}
    \R(\T \x; \bTheta) = \fr(\T \x;\bTheta) \;,
\end{align}
while also solving for geometric transformation parameters as $\T = [\bR|\bt] \in \SE(2)$ or $\T \in \SL(3)$, where $\bR \in \SO(2)$ and $\bt \in \Real^2$ denote the rigid rotation and translation, and $\T \in \SL(3)$ denotes the homography transformation matrix, respectively. We use another MLP with weights $\bPhi$ to model the coordinate-based warp neural field $\fw: \Real^2 \to \Real^3$ condition on the frame-dependent embedding $\bell$:
\begin{align} \label{eq:W2D}
    \W(\x;\bell,\bPhi) = \exp\big(\fw(\x;\bell,\bPhi)\big) \;,
\end{align}
where the operator $\exp(\cdot)$ denotes the exponential map from Lie algebra $\se(2)$ or $\lsl(3)$ to the Lie group $\SE(2)$ or $\SL(3)$, which ensures that the optimized transformation matrices $\T$ lie on the Lie group manifold during the gradient-based optimization.
\subsection{Bundle-Adjusting Neural Radiance Fields (3D)} \label{sec:nerf}

We then discuss the problem of \emph{simultaneously} recovering the 3D Neural Radiance Fields (NeRF)~\cite{mildenhall2020nerf} and the camera poses.
Given an 3D point, we predict the RGB color $\c \in \Real^3$ and volume density $\sigma \in \Real$ via an MLP $\fr: \Real^3 \to \Real^4$, which encodes the 3D scene using network parameters\footnotemark.
We begin by formulating NeRF's rendering process in the space of the camera view.
Denoting the homogeneous coordinates of pixel coordinates $\u \in \Real^2$ as $\x = [\u;1]^\top \in \Real^3$, the 3D point along the viewing ray at depth $z_i$ can be expressed as $z_i \x$, thus the query quantity $\y = [\c;\sigma]^\top = \fr(z_i \x;\bTheta)$, where $\bTheta$ is the parameters of $\fr$. Then the rendering color $\R$ at pixel location $\x$ can be composed by volume rendering
\begin{align} \label{eq:volrender}
    \R(\x) = \int_{z_\text{near}}^{z_\text{far}} T(\x,z) \sigma(z \x) \c(z \x) \mathrm{d}z \;,
\end{align}
where $T(\x,z) = \exp\!\big(\!-\!\int_{z_\text{near}}^{z} \sigma(z' \x) \mathrm{d}z' \big)$, and $z_\text{near}$ and $z_\text{far}$ are the near and far depth bounds of the scene.
Numerically, the integral formulation is discretely approximated using $K$ points sampled along a ray at depth $\{z_1,\dots,z_K\}$.
The network $\fr$ is evaluated $K$ times, and the outputs $\{\y_1,\dots,\y_K\}$ are then composited via volume rendering.
Denoting the differentiable and deterministic compositing function as $g: \Real^{4K} \to \Real^3$, such that $\R(\x)$ can be expressed as $\R(\x) = g\left(\y_1,\dots,\y_K\right)$.

\footnotetext{For the sake of simplicity, the viewing direction is omitted here.}

Here the camera poses are parametrized by $\T = [\bR|\bt] \in \SE(3)$, where $\bR \in \SO(3)$ and $\bt \in \Real^3$. 
Next, we use a 3D rigid transformation $\T$ to transform the 3D point $z_i \x$ from camera view space to world coordinates, and formulate the rendering color at pixel $\x$ as
\begin{align} \label{eq:comp2}
\R(\T \x; \bTheta) = g\Big(\fr(\T z_1\x;\bTheta),\dots,\fr(\T z_K\x;\bTheta)\Big).
\end{align}

Similar to neural image alignment, We use another MLP with weights $\bPhi$ to model the coordinate-based warp neural field $\fw: \Real^2 \to \Real^6$ condition on the frame-dependent embedding $\bell$:
\begin{align} \label{eq:W2D}
    \W(\x;\bell,\bPhi) = \exp\big(\fw(\x;\bell,\bPhi)\big) \;,
\end{align}
where the operator $\exp(\cdot)$ denotes the exponential map from Lie algebra $\se(3)$ to the Lie group $\SE(3)$.

\subsection{Differentiable Parameter Estimation} \label{sec:Estimation}

The local-to-global process allows L2G-NeRF to discover the correct registration with an initially flexible pixel-wise alignment and later shift focus to constrained parametric alignment. We derive the gradient flow of global alignment objective $\mathcal{L}^j_i = \big\|\T^j_i \x^j-\T^\ast_i \x^j\big\|_2^2$ \wrt the parameters $\bPhi$ of warp neural field $\W$ as
\begin{align}\small \label{eq:jacobian}
    \pderiv{\mathcal{L}^j_i}{\bPhi} = \pderiv{\mathcal{L}^j_i}{\T^j_i}\pderiv{\T^j_i}{\bPhi} + \pderiv{\mathcal{L}^j_i}{\T^\ast_i}\sum^{N}_{j=1}\pderiv{\T^\ast_i}{\T^j_i}\pderiv{\T^j_i}{\bPhi} \;.
\end{align}
Such that a differentiable solver is of critical importance to calculating the gradient of $\T^\ast_i$ \wrt $\T^j_i$ then backpropagated to update the parameters $\bPhi$. Next, we elaborate two differentiable solvers for rigid and homography transformation, respectively.

\noindent\textbf{Rigid parametric alignment.}
In the rigid parametric alignment problem, we assume $\{\T^j \x^j\}_{j=1}^N$ is transformed from $\{\x^j\}_{j=1}^N$ by an unknown global rigid transformation $\T = [\bR|\bt] \in \SE(2)$ or $\T = [\bR|\bt] \in \SE(3)$. To solve this classic orthogonal Procrustes problem~\cite{hurley1962procrustes}, we define centroids of $\{\x^j\}_{j=1}^N$ and $\{\T^j \x^j\}_{j=1}^N$ as
\begin{align}\label{eq:mean}
\overline{\x} = \frac{1}{N} \sum^{N}_{j=1} (\x^j)
\quad \textrm{and} \quad
\overline{\T\x} = \frac{1}{N} \sum^{N}_{j=1} (\T^j \x^j).
\end{align}
Then the cross-covariance matrix $\bH$ is given by
\begin{align}\label{eq:cross_covariance}
\bH = \sum_{j=1}^N(\x^j-\overline{\x})(\T^j \x^j-\overline{\T\x})^\top.
\end{align}
We use Singular Value Decomposition (SVD) to decompose $\bH$ as introduced in~\cite{umeyama1991least,kabsch1976solution}:
\begin{align}\label{eq:SVD}
\bH=\bU\bS\bV^\top.
\end{align}
Thus the optimal transformation minimizing Eq.~\eqref{eq:argmin} is given in closed form by 
\begin{align}\label{eq:close_form}
\bR = \bV\bU^\top\quad\textrm{and}\quad
\bt = -\bR\overline{\x}+\overline{\T\x}.
\end{align}

\noindent\textbf{Homography parametric alignment.}
In the homography parametric alignment problem, we assume $\{{\x^j}^{\prime}=\T^j \x^j\}_{j=1}^N$ is transformed from $\{\x^j\}_{j=1}^N$ by an unknown homography transformation $\T \in \SL(3)$. Written element by element, in homogenous coordinates, we get the following constraint:

\begin{align}\label{eq:element}
\left[\begin{array}{c}
\wu \\
\wv \\
1
\end{array}\right]=\left[\begin{array}{lll}
\T_{11} & \T_{12} & \T_{13} \\
\T_{21} & \T_{22} & \T_{23} \\
\T_{31} & \T_{32} & \T_{33}
\end{array}\right]\left[\begin{array}{c}
\ou \\
\ov \\
1
\end{array}\right].
\end{align}
Rearranging Eq.~\eqref{eq:element} as~\cite{abdel2015direct}, we get $\mathbf{A}^j \mathbf{h}=\mathbf{0}$, where
\begin{align}\small\label{eq:Rearranging}
\begin{aligned}
\mathbf{A}^j &=\left[\setlength{\arraycolsep}{1.0pt}\begin{array}{ccccccccc}
0 & 0 & 0 & -\ou & -\ov & -1 & \wv \ou & \wv \ov & \wv \\
\ou & \ov & 1 & 0 & 0 & 0 & -\wu \ou & -\wu \ov & -\wu
\end{array}\right] \\
\mathbf{h} &=\left(\T_{11}, \T_{12}, \T_{13}, \T_{21}, \T_{22}, \T_{23}, \T_{31}, \T_{32}, \T_{33}\right)^\top
\end{aligned}
\end{align}
Given the set of correspondences, we can form the linear system of equations $\bA \mathbf{h}=\mathbf{0}$, where $\bA =(\mathbf{A}^1 \dots \mathbf{A}^N)^{\top}$.
Thus we can solve the Homogeneous Linear Least Squares problem and calculate the non-trivial solution by SVD decomposition:
\begin{align}\small\label{eq:SVD2}
\bA=\bU\bS\bV^\top=\sum_{l=1}^9 \sigma_l \boldsymbol{u}_l \boldsymbol{v}_l^{\top},
\end{align}
where singular value $\sigma_l$ represents the reprojection error. Then we take the singular vector $\boldsymbol{v}_9$ that corresponds to the smallest singular value $\sigma_9$ as the solution of $\mathbf{h}$, and reshape it into the homography transformation matrix $\T$.

\begin{table*}[t!]
    \centering
    \begin{minipage}{0.415\linewidth}
        \includegraphics[width=1\linewidth,page=1]{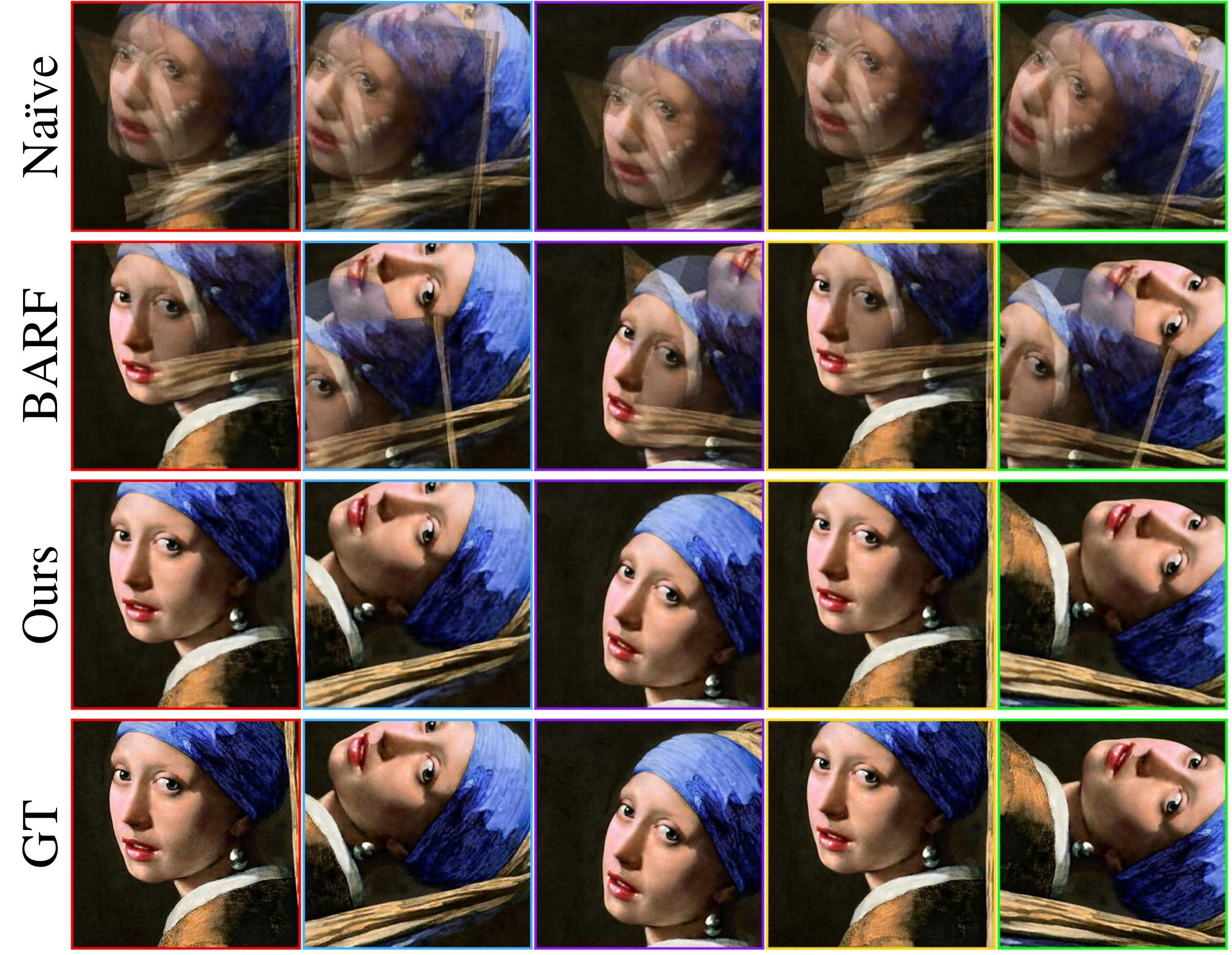}
        \vspace{-12pt}
        \captionof{figure}{
            Color-coded patch reconstructions of neural image alignment 
            under {\bf rigid} perturbations. The optimized warps are shown in Fig.~\ref{fig:rigid-results} with corresponding colors. 
            L2G-NeRF is able to recover accurate alignment and 
            photorealistic image reconstruction with local-to-global registration, while baselines result in suboptimal alignment.
        }
        \label{fig:rigid}
        
        \vspace{10pt}
        \includegraphics[width=1\linewidth,page=1]{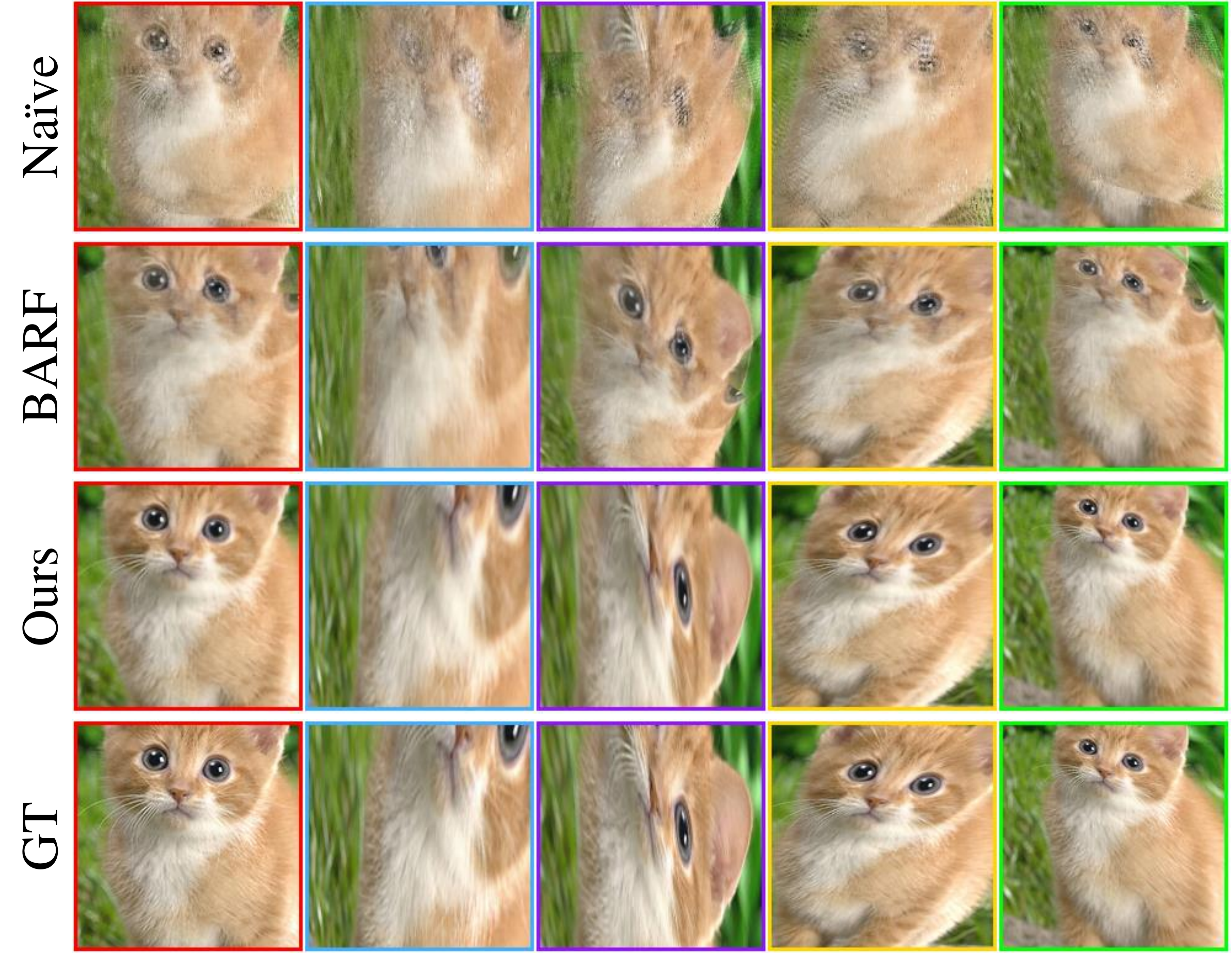}
        \vspace{-12pt}
        \captionof{figure}{
            Color-coded patch reconstructions of neural image alignment under {\bf homography} perturbations. 
        }
        \vspace{-8pt}
        \label{fig:homography}
    \end{minipage}
    \hspace{8pt}
    \begin{minipage}{0.555\linewidth}
        \includegraphics[width=1\linewidth,page=1]{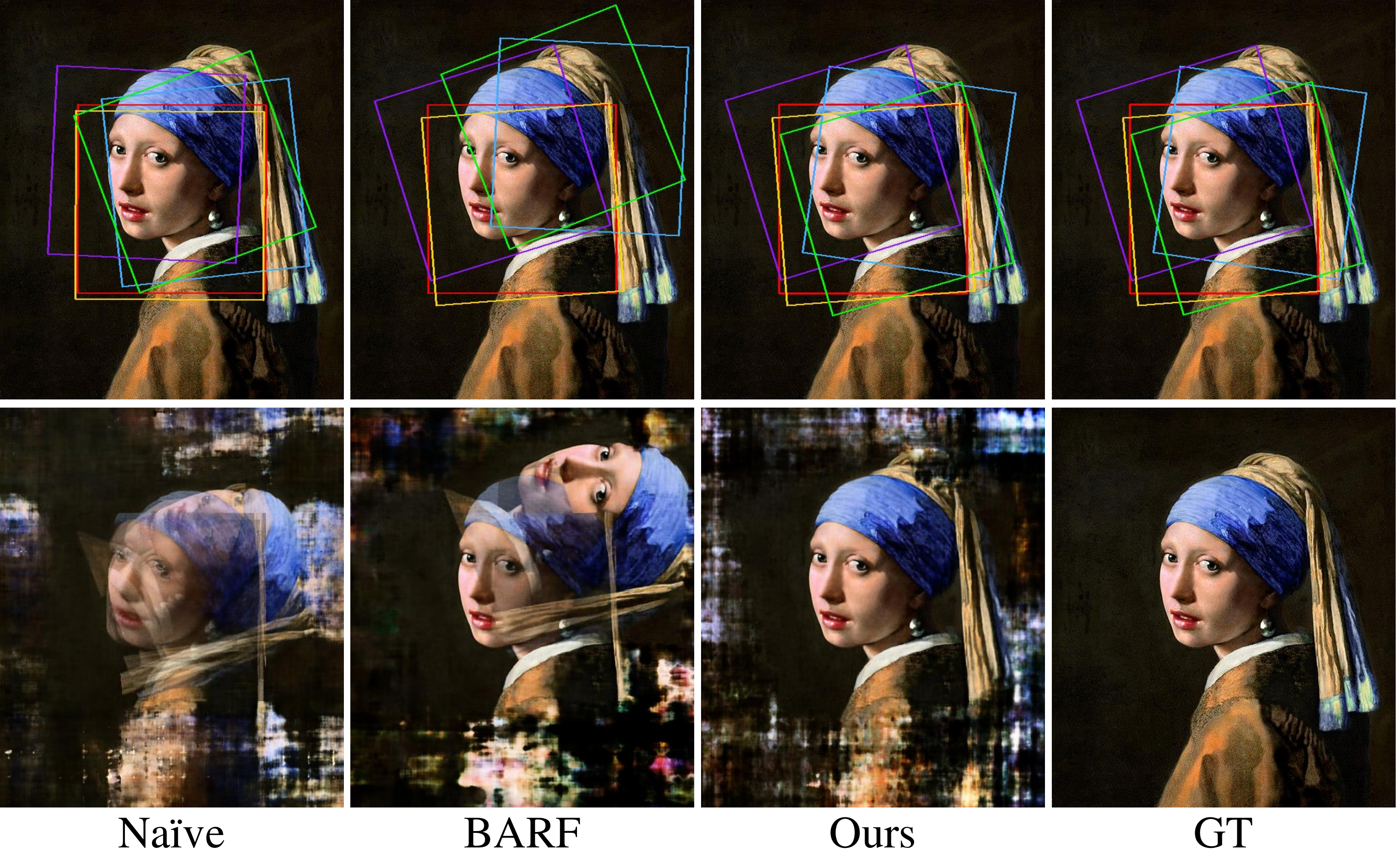}
        \vspace{-18pt}
        \captionof{figure}{
            Qualitative results of neural image alignment experiment under {\bf rigid} perturbations.
            Given color-coded patches (Fig.~\ref{fig:rigid}), we recover the alignment (top row) \emph{and} the neural field of the entire image (bottom row).
        }
        \label{fig:rigid-results}
        \vspace{4pt}
        \centering
        \resizebox{\linewidth}{!}{
            \begin{tabular}{c||c|c||c|c}
                \toprule
                \multirow{2}{*}{Method} & \multicolumn{2}{c||}{Rigid perturbations} & \multicolumn{2}{c}{Homography perturbations} \vspace{1.5pt} \\ 
                & Corner error (pixels) $\downarrow$ & Patch PSNR $\uparrow$ & Corner error (pixels) $\downarrow$ & Patch PSNR $\uparrow$ \\
                \midrule
                Na\"ive & 120.00 & 14.83 & 55.80 & 21.79 \\
                BARF~\cite{lin2021barf} & 110.20 & 17.78 & 30.21 & 23.24 \\
                Ours & 0.31 & 29.25 & 0.76 & 31.93 \\
                \bottomrule 
            \end{tabular}
        }
        \vspace{-8pt}
        \captionof{table}{
            Quantitative results of neural image alignment experiment under {\bf rigid} and {\bf homography} perturbations.
            L2G-NeRF optimizes for high-quality alignment and patch reconstruction, while baselines exhibit large errors.
        }
        \label{table:planar}
        \vspace{4pt}
        \includegraphics[width=1\linewidth,page=1]{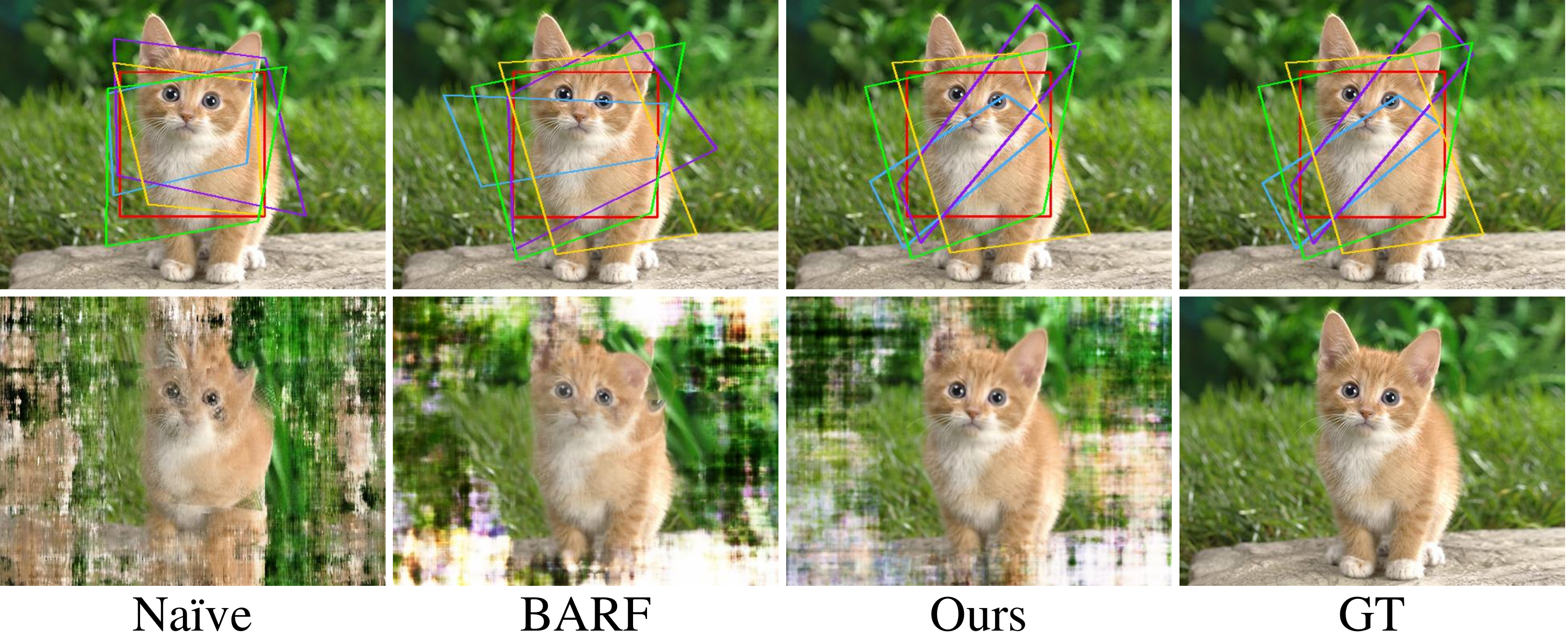}
        \vspace{-18pt}
        \captionof{figure}{
            Qualitative results of 
            neural image alignment experiment 
            under {\bf homography} perturbations.
        }
        \vspace{-8pt}
        \label{fig:homography-results}
    \end{minipage}
\end{table*}

\section{Experiments}
We first unfold the validation of L2G-NeRF and baselines on a 2D neural image alignment experiment, and then show that the local-to-global registration strategy can also be generalized to learn 3D neural fields (NeRF~\cite{mildenhall2020nerf}) from both synthetic data and photo collections.

\subsection{Neural image Alignment (2D)} \label{sec:exp-image}
We choose two representative images of ``Girl With a Pearl Earring'' renovation \copyright Koorosh Orooj \href{http://profoundism.com/free_licenses.html}{(CC BY-SA 4.0)}
and ``cat'' from ImageNet~\cite{deng2009imagenet} for rigid and homography image alignment experiments, respectively. As shown in Fig.~\ref{fig:rigid} and Fig.~\ref{fig:homography}, given $M=5$ patches sampled from the original image with rigid or homography perturbations, we optimize Eq.~\eqref{eq:soft_photometric_loss} to find the rigid transformation $\T \in \SE(2)$ or homography transformation $\T \in \SL(3)$ for each patch with network $\fw$, and learn the neural field of the entire image (Fig.~\ref{fig:rigid-results} and Fig.~\ref{fig:homography-results}) with network $\fr$ at the same time. 
We follow~\cite{lin2021barf} to initialize patch warps as identity and anchor the first warp to align the neural image to the raw image.

\begin{figure*}[t!]
    \centering  
    \includegraphics[width=0.98\linewidth,page=1]{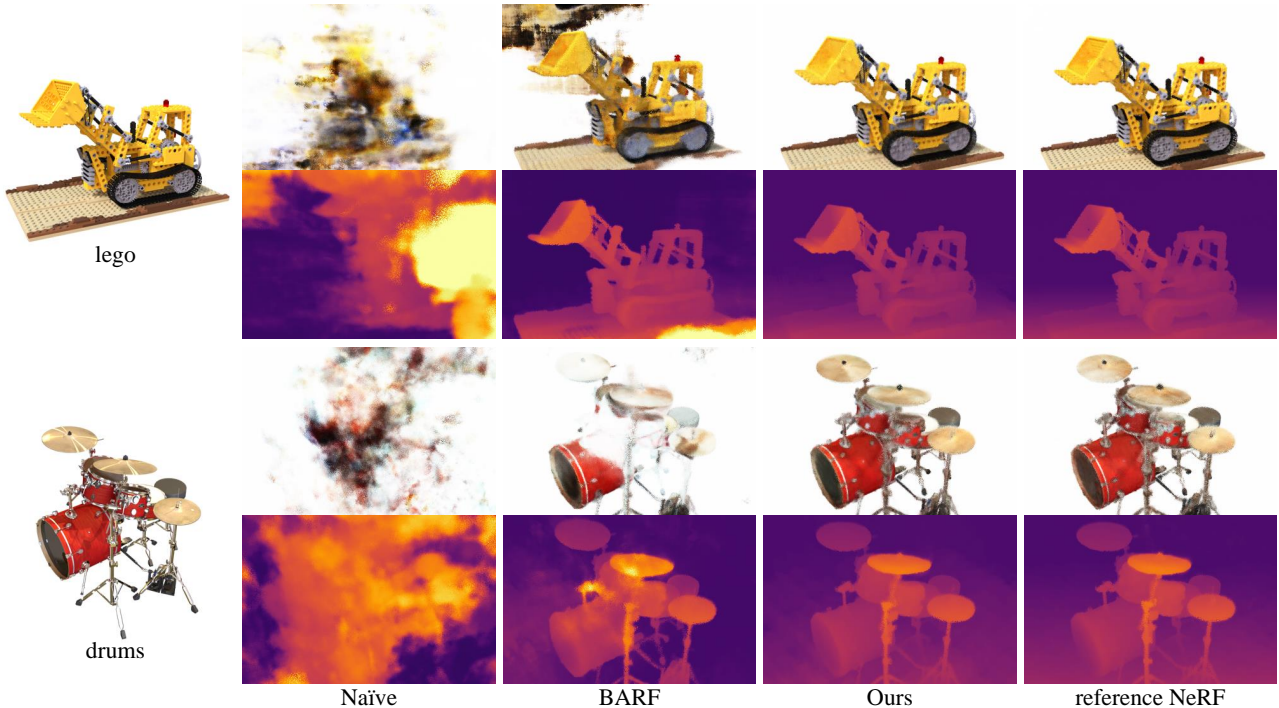}
    \vspace{-4pt}
    \caption{
        Qualitative results of bundle-adjusting neural radiance fields on synthetic scenes.
        The image synthesis and the expected depth are visualized with ray compositing in the top and bottom rows, respectively.
        While baselines render artifacts due to less-than-optimal registration, L2G-NeRF achieves qualified visual quality, which is comparable to the reference NeRF trained under ground-truth poses.
    }
    \label{fig:nerf-results-blender}
    \vspace{-8pt}
\end{figure*}

\noindent\textbf{Experimental settings.}
We evaluate our proposed method against a bundle-adjusting extension of the na\"ive 2D neural field, dubbed as Na\"ive, and the current state-of-the-art BARF~\cite{lin2021barf}, which employs a coarse-to-fine strategy for registration. We use the default coarse-to-fine scheduling, architecture and training procedure of neural field $\fr$ for both BARF and L2G-NeRF. 
For L2G-NeRF, We use a ReLU MLP for $\fw$ with six 256-dimensional hidden units, and
use the embedding with 128 dimensions for each image to model the frame-dependent embeddings $\{\bell_i\}_{i=1}^{M}$. We set multiplier $\lambda$ of the global alignment objective to $1\times10^{2}$.

\noindent\textbf{Results.}
We visualize the rigid and homography registration results in Fig.~\ref{fig:rigid-results} and Fig.~\ref{fig:homography-results}.
Alignment with Na\"ive results in ghosting artifacts in the recovered neural image due to large misalignment. On the other hand, alignment with BARF improves registration results but still falls into the suboptimal solutions, and struggles with image reconstruction. As L2G-NeRF discovers the precise geometric warps of all patches, it can optimize the neural image with high fidelity. We report the quantitative results in Table~\ref{table:planar}, where we use the mean average corner error (L2 distance between the ground truth corner position and the estimated corner position)~\cite{detone2016deep,le2020deep} and PSNR as the evaluation criteria for registration and reconstruction, respectively.
The experiment of image alignment shows how local-to-global strategy has a wide range of benefits for both rigid and homography registration for 2D neural fields, which can be easily extended to other geometric transformations. 

\begin{table*}[t!]
    \centering
    \setlength\tabcolsep{3pt}
    \resizebox{\linewidth}{!}{
        \begin{tabular}{c||ccc|ccc||ccc|c|ccc|c|ccc|c}
            \toprule
            \multirow{4}{*}{Scene} & \multicolumn{6}{c||}{Camera pose registration} & \multicolumn{12}{c}{View synthesis quality} 

            \\
            & \multicolumn{3}{c|}{Rotation ($\degree$) $\downarrow$} & \multicolumn{3}{c||}{Translation $\downarrow$} & \multicolumn{4}{c|}{PSNR $\uparrow$} & \multicolumn{4}{c|}{SSIM $\uparrow$} & \multicolumn{4}{c}{LPIPS $\downarrow$} \\
            \cmidrule{2-19}
            & \multirow{2}{*}{Na\"ive}
            & \multirow{2}{*}{BARF} 
            & \multirow{2}{*}{Ours}
            & \multirow{2}{*}{Na\"ive} 
            & \multirow{2}{*}{BARF} 
            & \multirow{2}{*}{Ours}
            & \multirow{2}{*}{Na\"ive} 
            & \multirow{2}{*}{BARF} 
            & \multirow{2}{*}{Ours} 
            & ref.
            & \multirow{2}{*}{Na\"ive} 
            & \multirow{2}{*}{BARF} 
            & \multirow{2}{*}{Ours} 
            & ref.
            & \multirow{2}{*}{Na\"ive} 
            & \multirow{2}{*}{BARF} 
            & \multirow{2}{*}{Ours} 
            & ref. \vspace{-2.5pt} \\
            & & &
            & & &
            & & & & NeRF
            & & & & NeRF
            & & & & NeRF \\
            
            \midrule
            Chair &  1.39 & 2.58 & \bf 0.14 & 60.32 & 10.43 & \bf 0.28 & 14.13 & 27.84 & \bf 30.99 & 31.93 & 0.83 & 0.92 & \bf 0.95 & 0.96 & 0.39 & 0.06 & \bf 0.05 & 0.04 \\
            Drums      &  7.99 & 4.54 & \bf 0.06 &  78.20 & 19.19 & \bf 0.40 & 11.63 & 21.92 & \bf 23.75 & 23.98 & 0.61 & 0.87 & \bf 0.90 & 0.90 & 0.62 & 0.14 & \bf 0.10 & 0.10 \\
            Ficus      &  3.13 & 1.65 & \bf 0.26 & 48.78 & 5.46 & \bf 1.11 & 14.30 & 25.85 & \bf 26.11 & 26.66 & 0.83 & \bf 0.93 & \bf 0.93 & 0.94 & 0.33 & 0.07 & \bf 0.06 & 0.05 \\
            Hotdog     &  7.04  & 2.42  & \bf{0.27} & 58.37  & 14.98 & \bf{1.42} & 15.10 & 27.34 & \bf{34.56} & 34.90  & 0.74 & 0.93 & \bf{0.97} & 0.97 & 0.42 & 0.06 & \bf{0.03} & 0.03 \\
            Lego       &  7.82  & 9.93  & \bf{0.09} & 81.93  & 47.42 & \bf{0.37} & 11.36 & 14.48 & \bf{27.71} & 29.29 & 0.61 & 0.69 & \bf{0.91} & 0.94 & 0.56 & 0.29 & \bf{0.06} & 0.04 \\
            Materials  &  5.57  & 0.68  & \bf{0.06} & 47.56  & 4.97  & \bf{0.28} & 11.51 & 26.29 & \bf{27.60} & 28.54 & 0.64 & 0.92 & \bf{0.93} & 0.94 & 0.49 & 0.08 & \bf{0.06} & 0.05 \\
            Mic        & 4.43  & 10.44 & \bf{0.10} & 77.47  & 45.66 & \bf{0.44} & 13.14 & 12.20 & \bf{30.91} & 31.96 & 0.85 & 0.76 & \bf{0.97} & 0.97 & 0.43 & 0.41 & \bf{0.05} & 0.04 \\
            Ship       &  11.10 & 23.90 & \bf{0.19} & 112.01 & 90.62 & \bf{0.61} & 9.41  & 8.19  & \bf{27.31} & 28.06 & 0.50 & 0.50 & \bf{0.85} & 0.86 & 0.64 & 0.63 & \bf{0.13} & 0.12 \\
            \midrule
            Mean       &  6.06  & 7.02  & \bf{0.15} & 70.58  & 29.84 & \bf{0.61} & 12.57 & 20.51 & \bf{28.62} & 29.42 & 0.70 & 0.82 & \bf{0.93} & 0.94 & 0.49 & 0.22 & \bf{0.07} & 0.06 \\
            \bottomrule
        \end{tabular}
    }
    \caption{
        Quantitative results of bundle-adjusting neural radiance fields on synthetic scenes.
        L2G-NeRF successfully optimizes camera poses, thus rendering high-quality images comparable to the reference NeRF model (trained using ground-truth camera poses), outperforming the baselines on all evaluation criteria.
        Translation errors are scaled by $100$.
    }
    \label{table:nerf-blender}
\end{table*}

\subsection{NeRF (3D): Synthetic Objects} \label{sec:exp-nerf-blender}

\begin{figure*}[t!]
    \centering  
    \includegraphics[width=1\linewidth,page=1]{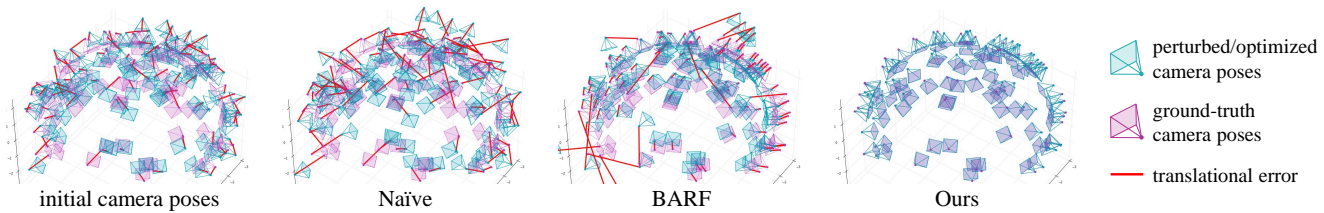}
    \vspace{-16pt}
    \caption{
        Visual comparison of the initial and optimized camera poses (Procrustes aligned) for the \textit{lego} scene.
        L2G-NeRF properly aligns all of the camera frames while baselines get stuck at suboptimal poses.
    }
    \label{fig:nerf-camera-blender}
    \vspace{-8pt}
\end{figure*}

This section investigates the challenge of learning 3D Neural Radiance Fields (NeRF)~\cite{mildenhall2020nerf} from noisy camera poses.
We evaluate L2G-NeRF and baselines on 8 synthetic object-centric scenes~\cite{mildenhall2020nerf}, in which each scene has $M=100$ rendered images with ground-truth camera poses for training.

\noindent\textbf{Experimental settings.}
For each scene, we synthetically perturb the camera poses $\T \in \SE(3)$ with additive noise $\xi \in \se(3)$ and $\xi \sim \mathcal{N}(\0,n\eye)$ as initial poses, where the multiplier $n$ is scene-dependent and given in the supplementary materials.
We assume known camera intrinsics and minimize the objective in Eq.~\eqref{eq:soft_photometric_loss} for optimizing the 3D neural fields $\fr$ and the warp field $\fw$ that finds rigid transformations relative to the initial poses. We evaluate L2G-NeRF against a na\"ive extension of the original NeRF model that jointly optimizes poses, dubbed as Na\"ive, and the coarse-to-fine bundle-adjusting neural radiance fields (BARF)~\cite{lin2021barf}.

\noindent\textbf{Implementation details.}
Our implementation of NeRF and BARF follows\cite{lin2021barf}.
For L2G-NeRF, We use a 6-layer ReLU MLP for $\fw$ with 256-dimensional hidden units.
We set multiplier $\lambda$ of the global alignment objective to $1\times10^{2}$ and employ the Adam optimizer to train all models for $200$K iterations with a learning rate that begins at $5\!\times\!10^{-4}$ for the 3D neural field $\fr$, and $1\!\times\!10^{-3}$ for the warp field $\fw$, and exponentially decays to $1\!\times\!10^{-4}$ and $1\!\times\!10^{-8}$, respectively. We follow the default coarse-to-fine scheduling for both BARF and L2G-NeRF.

\noindent\textbf{Evaluation criteria.}
Following BARF~\cite{lin2021barf}, we use Procrustes analysis to find a 3D similarity transformation that aligns the optimized poses to the ground truth before evaluating registration quality (quantitative results based on average translation and rotation errors), and perform test-time photometric pose optimization~\cite{lin2019photometric,yen2020inerf,lin2021barf} before evaluating view synthesis quality (quantitative results based on PSNR, SSIM and LPIPS~\cite{zhang2018unreasonable}).

\noindent\textbf{Results.}
We visualize the results in Fig.~\ref{fig:nerf-results-blender}, which are quantitatively reflected in Table~\ref{table:nerf-blender}.
On both sides of reconstruction and registration, L2G-NeRF achieves the best performance.
Fig.~\ref{fig:nerf-camera-blender} shows that L2G-NeRF can achieve near-perfect registration for the synthetic scenes.
Na\"ive NeRF suffers from suboptimal registration and ghosting artifacts.
BARF is able to recover a part of the pose misalignment and produce plausible reconstructions. However, it still suffers from blur artifacts like the fog effect around the objects. This fog effect is the consequence of BARF's attempt to reconstruct the scenes with half-baked registration.
We then compare the rendering quality to the reference standard NeRF (ref. NeRF), which is trained using ground truth poses, demonstrating that L2G-NeRF can achieve comparable image quality, despite being initialized from a significant camera pose misalignment.

\begin{figure*}[t!]
    \centering  
    \includegraphics[width=0.98\linewidth,page=1]{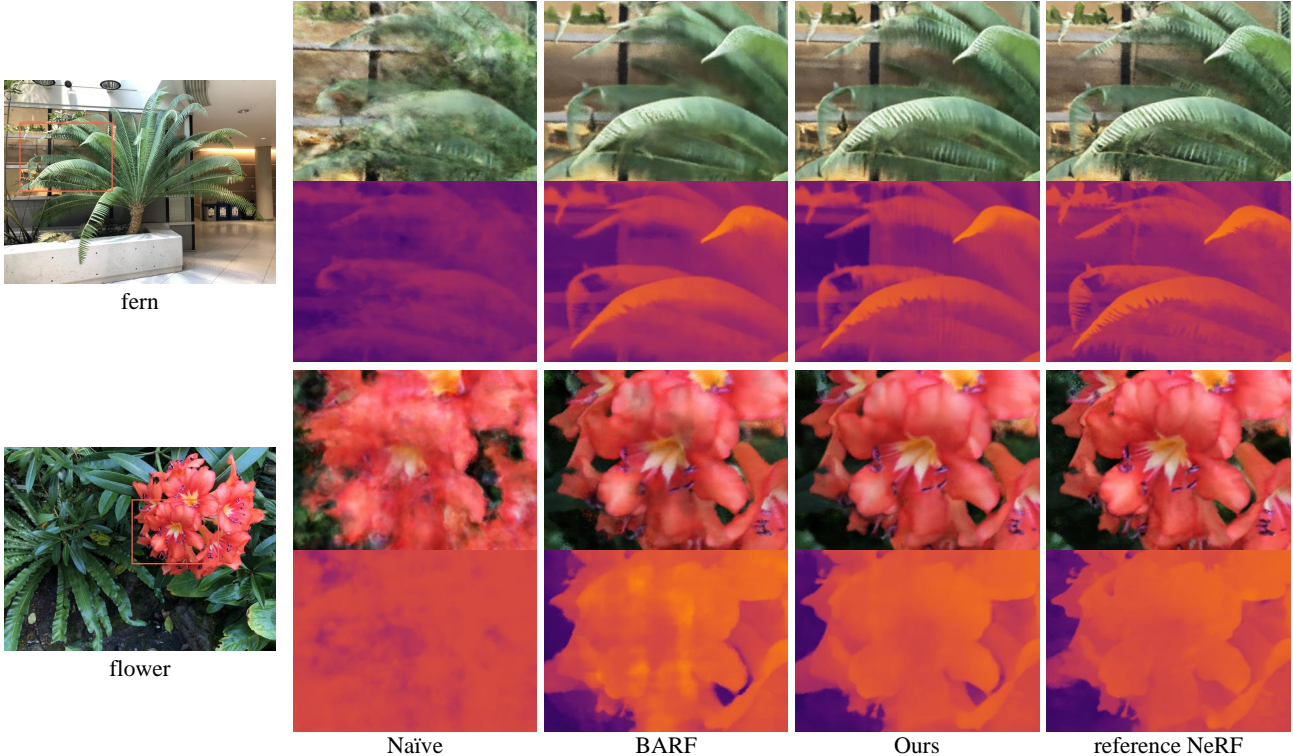}
    \vspace{-3pt}
    \caption{
        Qualitative results of bundle-adjusting neural radiance fields on real-world scenes. While BARF and L2G-NeRF can jointly optimize poses and scenes, L2G-NeRF produces higher fidelity results, which is competitive to reference NeRF trained under \SFM poses.
    }
    \label{fig:nerf-results-llff}
\end{figure*}
\begin{table*}[t!]
    \centering
    \setlength\tabcolsep{3pt}
    \resizebox{\linewidth}{!}{
        \begin{tabular}{c||ccc|ccc||ccc|c|ccc|c|ccc|c}
            \toprule
            \multirow{4}{*}{Scene} & \multicolumn{6}{c||}{Camera pose registration} & \multicolumn{12}{c}{View synthesis quality} 
            \\
            & \multicolumn{3}{c|}{Rotation ($\degree$) $\downarrow$} & \multicolumn{3}{c||}{Translation $\downarrow$} & \multicolumn{4}{c|}{PSNR $\uparrow$} & \multicolumn{4}{c|}{SSIM $\uparrow$} & \multicolumn{4}{c}{LPIPS $\downarrow$} \\
            \cmidrule{2-19}
            & \multirow{2}{*}{Na\"ive}
            & \multirow{2}{*}{BARF} 
            & \multirow{2}{*}{Ours}
            & \multirow{2}{*}{Na\"ive} 
            & \multirow{2}{*}{BARF} 
            & \multirow{2}{*}{Ours}
            & \multirow{2}{*}{Na\"ive} 
            & \multirow{2}{*}{BARF} 
            & \multirow{2}{*}{Ours} 
            & ref.
            & \multirow{2}{*}{Na\"ive} 
            & \multirow{2}{*}{BARF} 
            & \multirow{2}{*}{Ours} 
            & ref.
            & \multirow{2}{*}{Na\"ive} 
            & \multirow{2}{*}{BARF} 
            & \multirow{2}{*}{Ours} 
            & ref. \vspace{-2.5pt} \\
            & & &
            & & &
            & & & & NeRF
            & & & & NeRF
            & & & & NeRF \\
            
            \midrule
            Fern & 8.05 & \bf{0.17} & 0.20 & 1.74 & 0.19 & \bf{0.18} & 16.28 & 23.88 & \bf{24.57} & 24.19 & 0.39 & 0.71 & \bf{0.75} & 0.74 & 0.54 & 0.31 & \bf{0.26} & 0.25 \\
            Flower & 22.41  & \bf{0.31} & 0.33 & 5.81  & \bf{0.22} & 0.24 & 12.28 & 24.29 & \bf{24.90} & 22.97 & 0.21 & 0.71 & \bf{0.74} & 0.66 & 0.66 & 0.20 & \bf{0.17} & 0.26 \\
            Fortress & 171.77 & 0.41 & \bf{0.25} & 47.90 & 0.33 & \bf{0.25} & 11.56 & 29.06 & \bf{29.27} & 26.12 & 0.29 & 0.82 & \bf{0.84} & 0.79 & 0.83 & 0.13 & \bf{0.11} & 0.19 \\
            Horns & 29.42  & \bf{0.11} & 0.22 & 12.83 & \bf{0.16} & 0.27 & 8.94 & \bf{23.29} & 23.12 & 20.45 & 0.22 & \bf{0.74} & \bf{0.74} & 0.63 & 0.82 & 0.29 & \bf{0.26} & 0.41 \\
            Leaves & 79.47  & 1.13 & \bf{0.79} & 12.42 & \bf{0.24} & 0.34 & 9.10  & 18.91 & \bf{19.02} & 13.71 & 0.06 & 0.55 & \bf{0.56} & 0.21 & 0.80 & 0.35 & \bf{0.33} & 0.58 \\
            Orchids & 41.75  & \bf{0.60} & 0.67 & 19.99 & \bf{0.39} & 0.41 & 9.93 & 19.46 & \bf{19.71} & 17.26 & 0.09 & 0.57 & \bf{0.61} & 0.51 & 0.81 & 0.29 & \bf{0.25} & 0.31 \\
            Room & 175.06 & 0.31 & \bf{0.30} & 65.48 & 0.28 & \bf{0.23} & 11.48 & 32.05 & \bf{32.25} & 32.94 & 0.31 & 0.94 & \bf{0.95} & 0.95 & 0.85 & 0.10 & \bf{0.08} & 0.07 \\
            T-rex & 166.21 & 1.38 & \bf{0.89} & 55.02 & 0.86 & \bf{0.64} & 9.17 & 22.92 & \bf{23.49} & 21.86 & 0.16 & 0.78 & \bf{0.80} & 0.74 & 0.86 & 0.20 & \bf{0.16} & 0.25 \\
            \midrule
            Mean & 86.77 & 0.55 & \bf{0.46} & 27.65 & 0.33 & \bf{0.32} & 11.09 & 24.23 & \bf{24.54} & 22.44 & 0.22 & 0.73 & \bf{0.75} & 0.65 & 0.77 & 0.23 & \bf{0.20} & 0.29 \\
            \bottomrule 
        \end{tabular}
    }
    \vspace{-3pt}
    \caption{
        Quantitative results of bundle-adjusting neural radiance fields on real-world scenes.
        L2G-NeRF outperforms baselines and 
        achieves high-quality view synthesis that is competitive to reference NeRF trained under \SFM poses.
        Translation errors are scaled by $100$.
    }
    \vspace{-3pt}
    \label{table:nerf-llff}
\end{table*}

\subsection{NeRF (3D): Real-World Scenes} \label{sec:exp-nerf-llff}

We further explore the challenge of employing NeRF to learn 3D neural fields in real-world scenes with \emph{unknown} camera poses.
We evaluate our method and baselines on the standard benchmark LLFF dataset~\cite{mildenhall2019local}, which is captured by hand-held cameras that record 8 forward-facing scenes in the real world.

\noindent\textbf{Experimental settings.}
We initialize all cameras with the \emph{identity} transformation, \ie $\T_i=\eye \;\; \forall i$, and use the camera intrinsics provided by LLFF dataset.
We compare against the Na\"ive extension of NeRF~\cite{mildenhall2020nerf}, BARF~\cite{lin2021barf}, and use the same evaluation metrics as described in the experiments of synthetic objects (Sec.~\ref{sec:exp-nerf-blender}).

\noindent\textbf{Implementation details.}
We follow the same architectural settings and coarse-to-fine scheduling from the BARF~\cite{lin2021barf}.
For simplicity, We train without additional hierarchical sampling. 
We train all models for $200$K iterations 
with a learning rate of $1\!\times\!10^{-3}$ for the 3D neural field $\fr$ decaying to $1\!\times\!10^{-4}$, and $3\!\times\!10^{-3}$ for the warp field $\fw$ decaying to $1\!\times\!10^{-8}$. We use the same architecture of the warp field for L2G-NeRF described in Sec.~\ref{sec:exp-nerf-blender}.

\noindent\textbf{Results.}
Quantitative results are summarized in Table \ref{table:nerf-llff}. Na\"ive NeRF diverges to wrong camera poses,
producing poor view synthesis that cannot compete with BARF. In contrast, L2G-NeRF achieves competitive registration errors compared to BARF while outperforming the others on all view synthesis criteria. Actually, we note that the camera poses provided in LLFF are also estimations from \SFM packages~\cite{schonberger2016structure}; therefore, the pose evaluation is a noisy indication. 
Based on the fact that more accurate registration yields more photorealistic view synthesis, we recommend using view synthesis quality as the primary criterion for real-world scenes.
The high-fidelity visual quality shown in Fig.~\ref{fig:nerf-results-llff} highlights the ability of
L2G-NeRF to register cameras and reconstruct neural fields from scratch.
    
\section{Conclusion}

We present Local-to-Global Registration for Bundle-Adjusting Neural Radiance Fields (L2G-NeRF), which is demonstrated by extensive experiments that can effectively learn the neural fields of scenes and resolve large camera pose misalignment at the same time.
By establishing a unified formulation of bundle-adjusting neural fields, we demonstrate that local-to-global registration is beneficial for both 2D and 3D neural fields, allowing for various applications of diverse neural fields. Code and models will be made available to the research community to facilitate reproducible research.

Although local-to-global registration is much more robust than current state-of-the-art~\cite{lin2021barf}, L2G-NeRF still can not recover camera poses from scratch (identity transformation) for inward-facing $360\degree$ scenes, where large displacements of rotation exist. 
Specific methods such as epipolar geometry and graph optimization could be employed to handle these issues.

\renewcommand{\thesection}{\Alph{section}}
\setcounter{section}{0}
Here we provide more implementation details and experimental results. We encourage readers to view the supplementary video for an intuitive experience about different types of bundle-adjusting neural radiance fields.

\section{Additional Details}

\subsection{Time Consumption}

We implement all experiments on a single NVIDIA GeForce RTX 2080 Ti GPU. As shown in Table \ref{table:time}, L2G-NeRF takes about 4.5 and 8 hours for training in synthetic objects and real-world scenes, respectively, while training BARF~\cite{lin2021barf} takes about 8 and 10.5 hours. As a reference, we also compare time consumption against the ref. NeRF~\cite{mildenhall2020nerf} trained under ground-truth poses (without the requirement of optimizing poses), showing that L2G-NeRF can achieve comparable time consumption.
The time analysis indicates that calculating the gradient \wrt local pose (local-to-global registration) is more efficient than calculating the gradient \wrt global pose (global registration).

\begin{table}[t]
    \centering
    \setlength\tabcolsep{10pt}
    \resizebox{\linewidth}{!}{
        \begin{tabular}{c||ccc}
        \toprule
         Scene & BARF & Ours & ref. NeRF \\
        \midrule
         Synthetic objects   & 08:18  & 04:35  & 04:30 \\
         Real-World scenes   & 10:38   & 07:42   & 07:25 \\
        \bottomrule 
        \end{tabular}
    }
    \vspace{-6pt}
    \caption{
        Average training time (hh:mm).
    }
    \label{table:time}
\end{table}

\begin{table}[t]
    \centering
    \setlength\tabcolsep{2pt}
    \resizebox{\linewidth}{!}{
        \begin{tabular}{c||cccccccc}
        \toprule
        Scene & Chair & Drums & Ficus & Hotdog & Lego & Materials & Mic  & Ship \\
        \midrule
        $n_r$  & 0.01  & 0.05  & 0.03  & 0.04   & 0.07 & 0.04      & 0.04 & 0.09 \\
        $n_t$  & 0.4   & 0.5   & 0.3   & 0.4    & 0.5  & 0.3       & 0.5  & 0.7 \\
        \bottomrule 
        \end{tabular}
    }
    \vspace{-6pt}
    \caption{
        Multiplier of pose perturbation for synthetic scenes.
    }
    \label{table:blender_noise}
\end{table}

\begin{table}[t]
    \centering
    \setlength\tabcolsep{2pt}
    \resizebox{\linewidth}{!}{
        \begin{tabular}{c||cccccccc}
        \toprule
        Scene & Fern & Flower & Fortress & Horns & Leaves & Orchids & Room & T-rex \\
        \midrule
        $\lambda$ 
        & $1\hspace{-4pt}\times\hspace{-4pt}10^2$ 
        & $1\hspace{-4pt}\times\hspace{-4pt}10^3$ 
        & $1\hspace{-4pt}\times\hspace{-4pt}10^5$ 
        & $1\hspace{-4pt}\times\hspace{-4pt}10^5$ 
        & $1\hspace{-4pt}\times\hspace{-4pt}10^2$ 
        & $1\hspace{-4pt}\times\hspace{-4pt}10^2$ 
        & $1\hspace{-4pt}\times\hspace{-4pt}10^5$ 
        & $1\hspace{-4pt}\times\hspace{-4pt}10^5$ \\
        \bottomrule 
        \end{tabular}
    }
    \vspace{-6pt}
    \caption{
        Multiplier $\lambda$ of global alignment objective.
    }
    \label{table:lambda}
\end{table}

\subsection{Camera Pose Perturbation}

In all experiments, we always use the same initial conditions for all methods (fixed random seeds).
For each object of synthetic scenes, we perturb the camera poses with additive noise as initial poses. Note that the way we add noise differs from ~\cite{lin2021barf}, which perturbs ground-truth camera poses using left multiplication (transform cameras around the object's center). Transformed cameras almost still face the object's center, and the distances between the cameras and the object are almost unchanged. In contrast, we perturb ground-truth camera poses using right multiplication (transform cameras around themselves), thereby perturbing camera viewing directions (not always toward the object's center) and camera positions (including the distances from them to the object), respectively.

The 6-DoF perturbation is parametrized by $\T = [\bR|\bt] \in \SE(3)$, where $\bR \in \SO(3)$, $\bt \in \Real^3$, and $\bR$ is generated by exponential map $\exp\big(\mathbf{r}\big)$ from the Lie algebra $\so(3)$ to the Lie group $\SO(3)$.
The additive rotation noise $\mathbf{r} \in \so(3)$ and translation noise $\bt \in \Real^3$ are distributed as $\mathbf{r} \sim \mathcal{N}(\0,n_r\eye)$ and $\bt \sim \mathcal{N}(\0,n_t\eye)$, where the multiplier $n_r$ and $n_t$ are scene-dependent and given in Table \ref{table:blender_noise}. 

\begin{table}[t!]
    \centering
    \setlength\tabcolsep{1pt}
    \resizebox{\linewidth}{!}{
        \begin{tabular}{c||cccc|cccc||cccc|cccc}
            \toprule
            \multirow{3}{*}{Scene} & \multicolumn{8}{c||}{Camera pose registration} & \multicolumn{8}{c}{View synthesis quality} 
            \\
            & \multicolumn{4}{c|}{Rotation ($\degree$) $\downarrow$} & \multicolumn{4}{c||}{Translation $\downarrow$} & \multicolumn{4}{c|}{PSNR $\uparrow$} & \multicolumn{4}{c}{LPIPS $\downarrow$} \\
            \cmidrule{2-17}

            &$1\hspace{-4pt}\times\hspace{-4pt}10^2$  &$1\hspace{-4pt}\times\hspace{-4pt}10^3$  &$1\hspace{-4pt}\times\hspace{-4pt}10^4$  &$1\hspace{-4pt}\times\hspace{-4pt}10^5$  
            &$1\hspace{-4pt}\times\hspace{-4pt}10^2$  &$1\hspace{-4pt}\times\hspace{-4pt}10^3$  &$1\hspace{-4pt}\times\hspace{-4pt}10^4$  &$1\hspace{-4pt}\times\hspace{-4pt}10^5$  
            &$1\hspace{-4pt}\times\hspace{-4pt}10^2$  &$1\hspace{-4pt}\times\hspace{-4pt}10^3$  &$1\hspace{-4pt}\times\hspace{-4pt}10^4$  &$1\hspace{-4pt}\times\hspace{-4pt}10^5$  
            &$1\hspace{-4pt}\times\hspace{-4pt}10^2$  &$1\hspace{-4pt}\times\hspace{-4pt}10^3$  &$1\hspace{-4pt}\times\hspace{-4pt}10^4$  &$1\hspace{-4pt}\times\hspace{-4pt}10^5$ \\
            \midrule
            Flower 
            &0.44 	&\bf{0.33} 	&\textbackslash{}	&\textbackslash{}
            &0.30 	&\bf{0.24} 	&\textbackslash{}	&\textbackslash{}
            &24.59	&\bf{24.90} 	&\textbackslash{}	&\textbackslash{}
            &0.18	&\bf{0.17} 	&\textbackslash{}	&\textbackslash{} \\
            Horns 
            &0.36 	&0.24 	&0.23 	&\bf{0.22} 
            &0.80 	&0.57 	&0.32 	&\bf{0.27}
            &22.51 	&22.84 	&22.82 	&\bf{23.12}
            &0.28	&0.28 	&0.27 	&\bf{0.26} \\
            \bottomrule 
        \end{tabular}
    }
    \vspace{-6pt}
    \caption{
        Ablation on the global alignment objective multiplier $\lambda$.
    }
    \label{table:multiplier}
\end{table}

\begin{figure}[t]
\centering
  \includegraphics[width=1.0\linewidth]{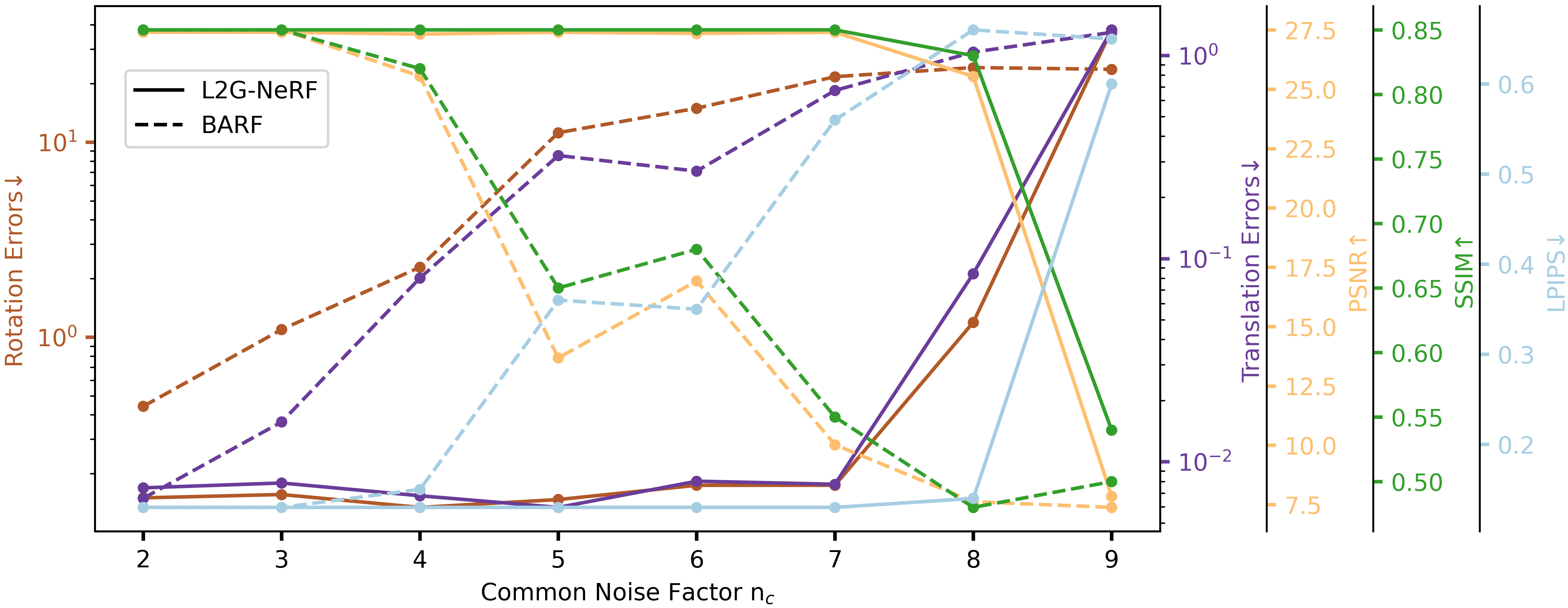}
    \vspace{-8pt}
  \caption{Convergence \wrt camera pose perturbation.}
\vspace{-6pt}
\label{fig:Convergence}
\end{figure}

\begin{table*}[t!]
    \centering
    \begin{minipage}{0.415\linewidth}
        \includegraphics[width=1\linewidth,page=1]{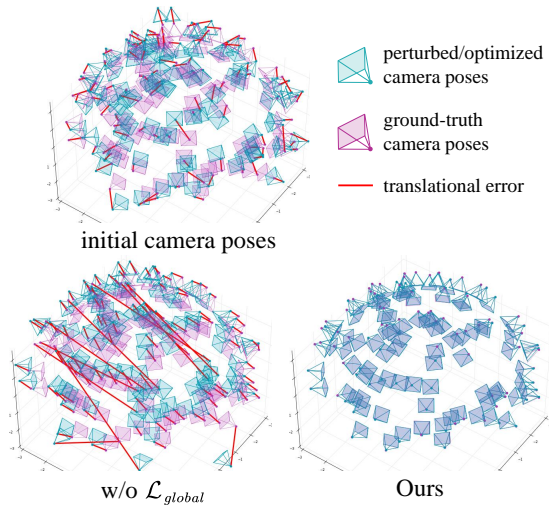}
        \vspace{-6pt}
        \captionof{figure}{
            Visual comparison of ablation study about optimized camera poses (Procrustes aligned) for \textit{hotdog} object.
            Full L2G-NeRF successfully aligns camera frames while w/o $\mathcal{L}_{global}$ gets stuck at suboptimal poses.
        }
        \vspace{-4pt}
        \label{fig:ab_synthetic_camera}
    \end{minipage}
    \hspace{8pt}
    \begin{minipage}{0.555\linewidth}
        \includegraphics[width=1\linewidth,page=1]{figSupp/ab_synthetic.pdf}
        \vspace{-16pt}
        \captionof{figure}{
                Ablation study of NeRF on \textit{hotdog} synthetic object.
                The image synthesis and the expected depth are visualized with ray compositing in the top and bottom rows, respectively.
                Full L2G-NeRF achieves comparable rendering quality to the reference NeRF (trained using ground-truth poses), while ablation w/o $\mathcal{L}_{global}$ renders artifacts due to suboptimal registration.
        }
        \label{fig:ab_synthetic}
        \vspace{-4pt}
    \end{minipage}
\end{table*}

\begin{table*}[t!]
    \centering
    \begin{minipage}{0.415\linewidth}
        \includegraphics[width=1\linewidth,page=1]{figSupp/ab_llff_camera.pdf}
        \vspace{-16pt}
        \captionof{figure}{
            Visualization of ablation study about registration for {\it room} scene.
            Results from L2G-NeRF highly agree with \SFM
            ~\cite{schonberger2016structure} 
            (colored in black), whereas  w/o $\mathcal{L}_{global}$ results in suboptimal alignment.
        }
        \label{fig:ab_llff_camera}
        \vspace{-2pt}
    \end{minipage}
    \hspace{8pt}
    \begin{minipage}{0.555\linewidth}
        \includegraphics[width=1\linewidth,page=1]{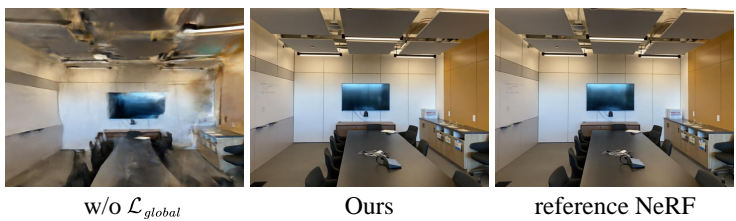}
        \vspace{-8pt}
        \captionof{figure}{
            Ablation study of NeRF on {\it room} real-world scenes from \emph{unknown} camera poses.
            While L2G-NeRF can jointly optimize poses and scenes, L2G-NeRF produces high fidelity results, which is competitive to reference NeRF trained using \SFM poses.
            Ablation w/o $\mathcal{L}_{global}$ diverges to wrong poses and hence produces ghosting artifacts.
        }
        \label{fig:ab_llff}
        \vspace{-2pt}
    \end{minipage}
\end{table*}

\begin{table*}[t!]
    \centering
    \setlength\tabcolsep{2pt}
    \resizebox{\linewidth}{!}{
        \begin{tabular}{c||ccc|ccc||ccc|c|ccc|c|ccc|c}
            \toprule
            \multirow{4}{*}{Scene} & \multicolumn{6}{c||}{Camera pose registration} & \multicolumn{12}{c}{View synthesis quality} \vspace{1.5pt} \\
            & \multicolumn{3}{c|}{Rotation ($\degree$) $\downarrow$} & \multicolumn{3}{c||}{Translation $\downarrow$} & \multicolumn{4}{c|}{PSNR $\uparrow$} & \multicolumn{4}{c|}{SSIM $\uparrow$} & \multicolumn{4}{c}{LPIPS $\downarrow$} \\
            \cmidrule{2-19}
            &  \small Global
            &  \small Local
            &  \small L2G
            &  \small Global
            &  \small Local
            &  \small L2G
            &  \small Global
            &  \small Local
            &  \small L2G 
            &  \small ref.
            &  \small Global
            &  \small Local
            &  \small L2G 
            &  \small ref.
            &  \small Global
            &  \small Local
            &  \small L2G 
            &  \small ref. \\
            &  \small BARF
            &  \small w/o $\mathcal{L}_{g}$ 
            &  \small Ours
            &  \small BARF
            &  \small w/o $\mathcal{L}_{g}$ 
            &  \small Ours
            &  \small BARF
            &  \small w/o $\mathcal{L}_{g}$ 
            &  \small Ours
            &  \small NeRF
            &  \small BARF
            &  \small w/o $\mathcal{L}_{g}$ 
            &  \small Ours
            &  \small NeRF
            &  \small BARF
            &  \small w/o $\mathcal{L}_{g}$ 
            &  \small Ours
            &  \small NeRF \\
            \midrule
            Synthetic objects & 7.02 & 3.63 & \textbf{0.15} & 29.84  & 14.34 & \textbf{0.61} & 20.51 & 22.70 & \textbf{28.62} & 29.42 & 0.82  & 0.85 & \textbf{0.93} & 0.94 & 0.22  & 0.14 & \textbf{0.07} & 0.06 \\
            Real-World scenes & 0.55 & 23.82 & \textbf{0.46} & 0.33 & 10.66 & \textbf{0.32} & 24.23 & 20.71 & \textbf{24.54} & 22.44 & 0.73 & 0.64 & \textbf{0.75} & 0.65 & 0.23 & 0.33 & \textbf{0.20} & 0.29 \\
            \bottomrule 
        \end{tabular}
    }
    \vspace{-4pt}
    \caption{
        Quantitative results of ablation study about bundle-adjusting neural radiance fields.
        L2G-NeRF outperforms the local registration method (ablation w/o $\mathcal{L}_{global}$) and global registration method (BARF)
        on the average evaluation criteria of both synthetic objects and real-world scenes, which reveals the advantage of our local-to-global registration process.
        Translation errors are scaled by $100$.
    }
    \vspace{-4pt}
    \label{table:nerf-ablation}
\end{table*}

\subsection{Convergence}

We analyze the convergence of joint optimization on the \textit{Ship} scene.
We first set the base rotation noise multiplier $n_r$ as $0.01$ and the base translation noise multiplier $n_t$ as $0.1$, then linearly increased them by a common factor of $\{n_c\}_{n_c=2}^{9}$.
As shown in \cref{fig:Convergence}, BARF fails to converge with $n_c$=$4$ ($n_r$=$0.04$,$ n_t$=$0.4$) while L2G-NeRF fails to converge with $n_c$=$8$ ($n_r$=$0.08$,$n_t$=$0.8$). Moreover, we also analyze the influence of individual noise. 
Let $n_r$=$0$, BARF and L2G-NeRF can handle the largest $n_t$ of $0.6$ and $1.1$, respectively. 
Let $n_t$=$0$, BARF and L2G-NeRF can handle the largest $n_r$ of $0.16$ and $0.25$, respectively.
In more noisy cases (such as random init), all methods cannot converge.

\begin{figure*}[t]
\centering
  \includegraphics[width=1\linewidth]{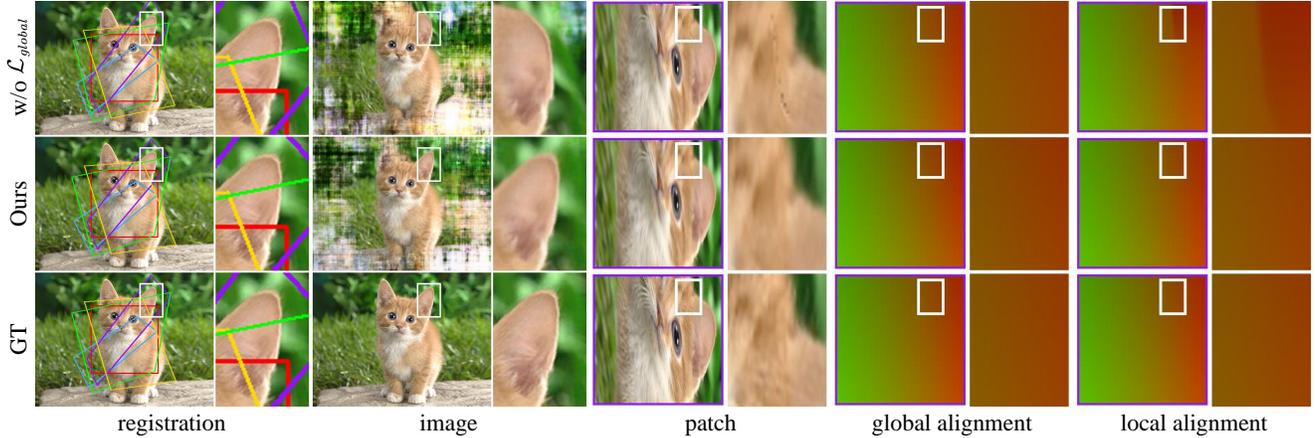}
    \vspace{-16pt}
  \caption{
  Ablation study of the neural image alignment experiment.
  Given color-coded image patches, we aim to recover the alignment and the neural field of the entire image.
  L2G-NeRF is able to find proper alignment and reconstruct high-fidelity neural image, while w/o $\mathcal{L}_{global}$ falls into false local alignments that do not obey the geometric constraint (global alignments), which results in ambiguous registration and distorted reconstruction (cat ears).
  }
    \vspace{-1pt}
\label{fig:ab_cat}
\end{figure*}

\begin{figure*}[t]
\centering
  \includegraphics[width=1\linewidth]{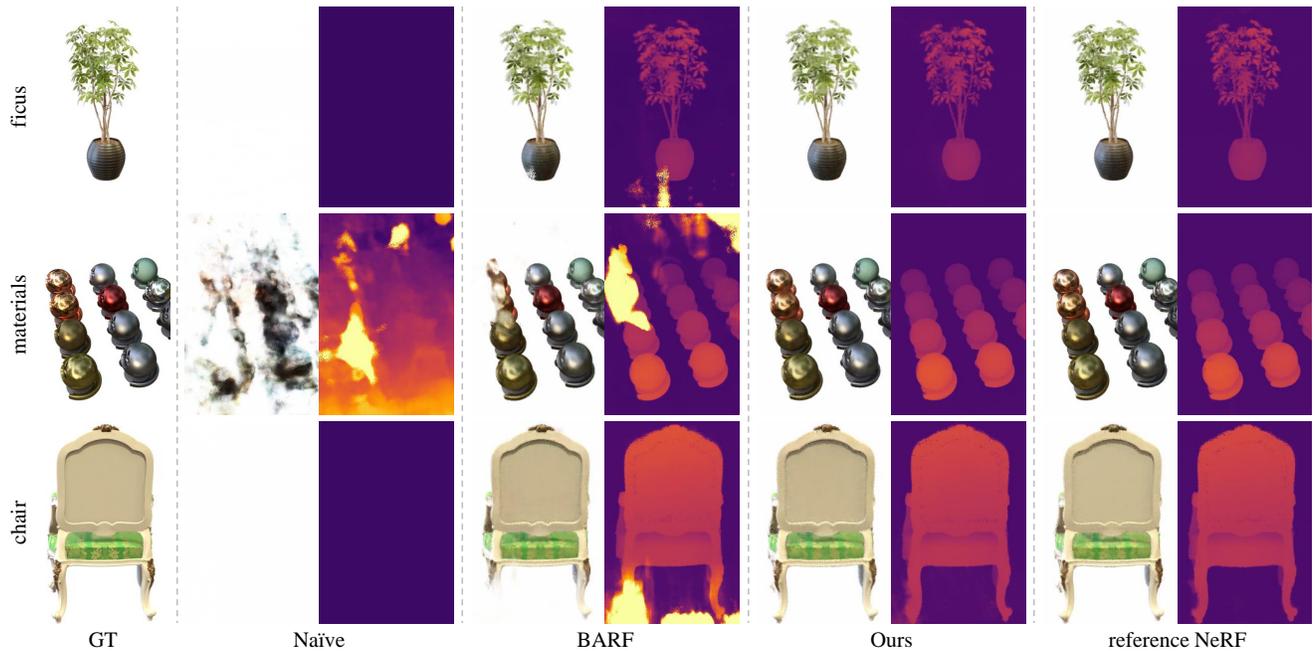}
    \vspace{-16pt}
  \caption{Additional qualitative results of bundle-adjusting neural radiance fields on synthetic scenes.
        The image synthesis and the expected depth are visualized with ray compositing in the left and right columns.
        While baselines render artifacts due to suboptimal solutions, L2G-NeRF achieves qualified visual quality, which is comparable to the reference NeRF trained using ground-truth poses.
        }
\label{fig:result_synthetic}
\end{figure*}

\subsection{Tuning Parameters}

We set the multiplier $\lambda$ of the global alignment objective to $1\times10^{2}$ for both the neural image alignment experiment and learning NeRF from imperfect camera poses with synthetic object-centric scenes. To further solve the challenging problem of learning NeRF in forward-facing LLFF scenes from \emph{unknown} poses, we float the multiplier $\lambda$ between $1\times10^{2}$ and $1\times10^{5}$ (summarized in Table \ref{table:lambda}) to achieve preferable results for specific scenes.
As shown in Table \ref{table:multiplier}, 
a larger $\lambda$ encourages the model to emphasize geometric constraints more, achieving better accuracy but worse robustness (fails to converge on the \textit{Flower} scene).

\begin{figure*}[t]
\centering
  \includegraphics[width=1\linewidth]{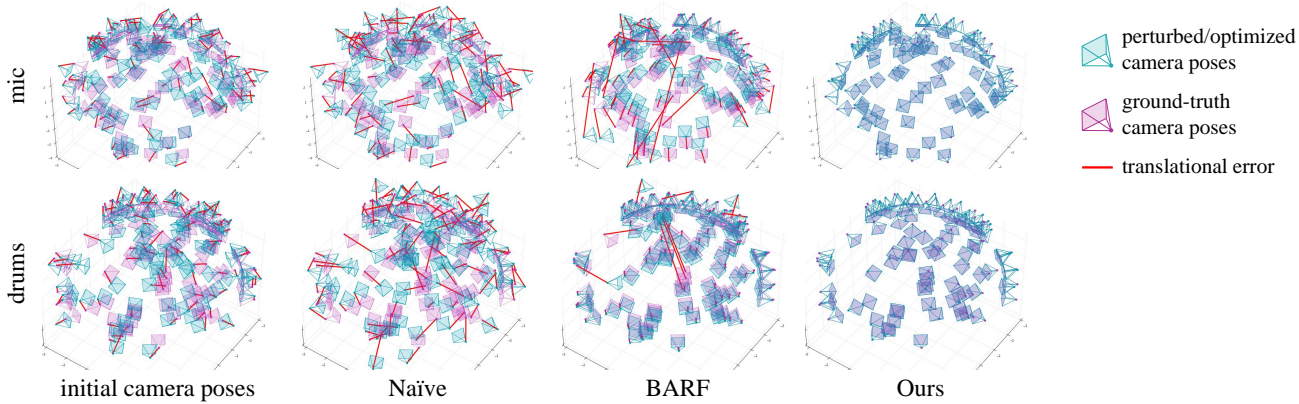}
    \vspace{-17pt}
  \caption{
        Additional visual comparison of the optimized camera poses (Procrustes aligned) for the \textit{mic} and \textit{drums} objects.
        L2G-NeRF successfully aligns all the camera frames while baselines get stuck at suboptimal solutions.
  }
\label{fig:result_synthetic_camera}
\end{figure*}

\begin{figure*}[t]
\centering
  \includegraphics[width=1\linewidth]{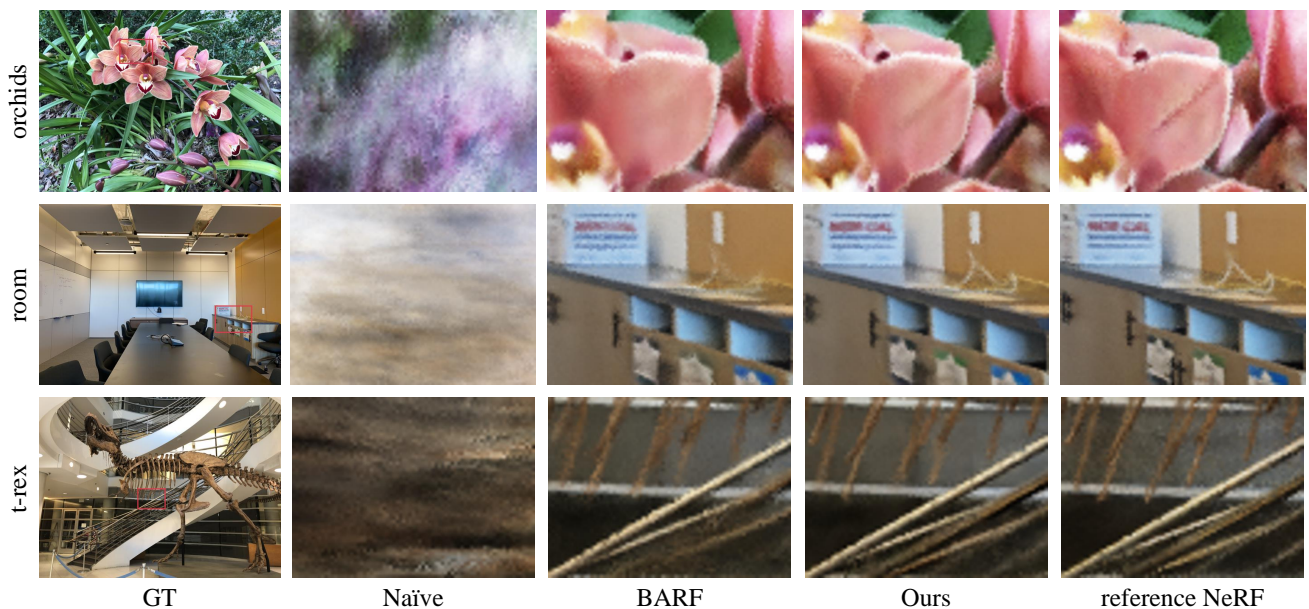}
  \vspace{-17pt}
  \caption{Additional novel view synthesis results of NeRF on real-world
  scenes (LLFF dataset) from \emph{unknown} camera poses.
  L2G-NeRF can optimize for neural fields of higher quality than baselines, while achieving the comparable quality of the reference NeRF model that is trained under the camera poses provided by \SFM~\cite{schonberger2016structure}.
  }
\label{fig:result_llff}
\end{figure*}

\begin{figure*}[t!]
\centering
  \includegraphics[width=1\linewidth]{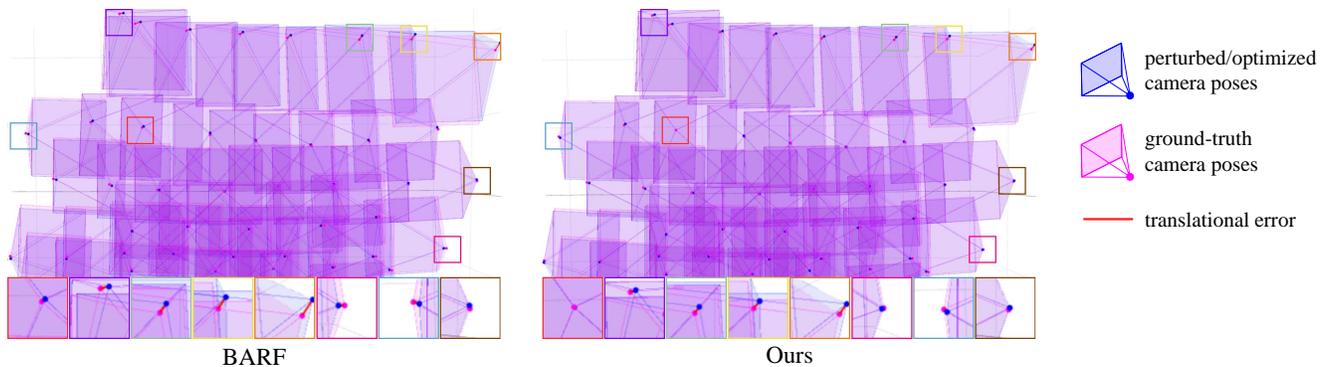}
    \vspace{-17pt}
  \caption{Visual comparison of the optimized camera poses (Procrustes aligned) for the \textit{t-rex} real-world scene.
  L2G-NeRF successfully recovers the camera poses from \emph{identity} transformation, which achieves fewer errors than BARF.}
\label{fig:result_llff_camera2}
\end{figure*}

\begin{table*}[t!]
    \centering
    \begin{minipage}{0.415\linewidth}
        \includegraphics[width=1\linewidth,page=1]{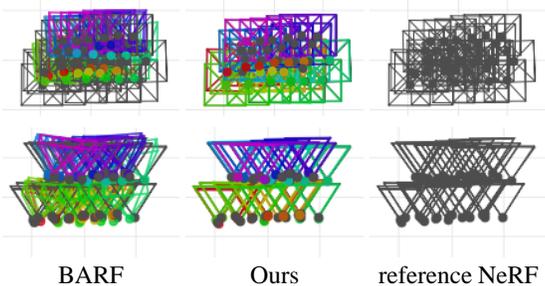}
        \vspace{-10pt}
        \captionof{figure}{
            Visual comparison of optimized camera poses (Procrustes aligned) for the challenging \textit{toys} scene captured under large displacements (hierarchical camera poses).
            L2G-NeRF successfully aligns all camera frames, which highly agrees with \SFM
            ~\cite{schonberger2016structure} camera poses
            (colored in black),
            while BARF gets stuck at suboptimal solutions.
        }
        \vspace{-4pt}
        \label{fig:iphone_camera1}
    \end{minipage}
    \hspace{8pt}
    \begin{minipage}{0.555\linewidth}
        \includegraphics[width=1\linewidth,page=1]{figSupp/nerf_toys.pdf}
        \vspace{-16pt}
        \captionof{figure}{
                Results of NeRF on \textit{toys} scene.
                L2G-NeRF achieves comparable synthesis quality to the reference NeRF (trained under \SFM camera poses). But BARF fails to recover the proper geometry, which results in artifacts.}
        \label{fig:iphone_nerf1}
        \vspace{-4pt}
    \end{minipage}
\end{table*}

\begin{table*}[t!]
    \centering
    \begin{minipage}{0.415\linewidth}
        \includegraphics[width=1\linewidth,page=1]{figSupp/camera_foods.pdf}
        \vspace{-6pt}
        \captionof{figure}{
            Visual comparison of optimized camera poses for the challenging \textit{foods} scene captured under sparse views.
            Results from L2G-NeRF highly agree with \SFM, whereas BARF results in suboptimal alignment.
        }
        \label{fig:iphone_camera2}
        \vspace{-1pt}
    \end{minipage}
    \hspace{8pt}
    \begin{minipage}{0.555\linewidth}
        \includegraphics[width=1\linewidth,page=1]{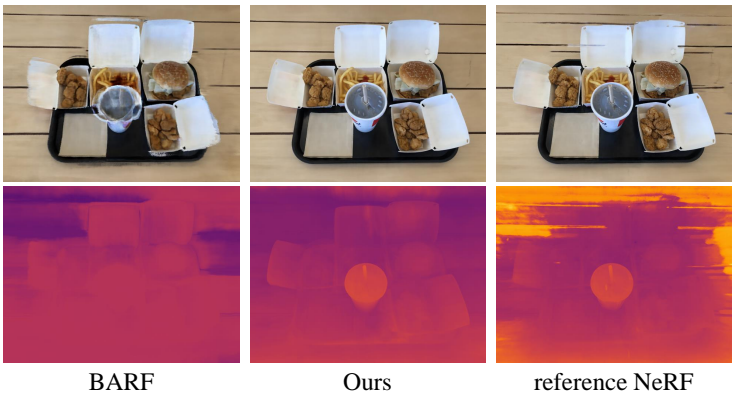}
        \vspace{-16pt}
        \captionof{figure}{
            Results of NeRF on \textit{foods} scene.
            L2G-NeRF outperforms BARF and even achieves better performance than reference NeRF in the scene where \SFM~\cite{schonberger2016structure} struggles with finding accurate registration from sparse views.
        }
        \label{fig:iphone_nerf2}
        \vspace{-1pt}
    \end{minipage}
\end{table*}

\begin{table*}[t!]
    \centering
    \setlength\tabcolsep{3pt}
    \resizebox{\linewidth}{!}{
        \begin{tabular}{c||ccc|ccc||ccc|c|ccc|c|ccc|c}
            \toprule
            \multirow{4}{*}{Scene} & \multicolumn{6}{c||}{Camera pose registration} & \multicolumn{12}{c}{View synthesis quality} 
            \\
            & \multicolumn{3}{c|}{Rotation ($\degree$) $\downarrow$} & \multicolumn{3}{c||}{Translation $\downarrow$} & \multicolumn{4}{c|}{PSNR $\uparrow$} & \multicolumn{4}{c|}{SSIM $\uparrow$} & \multicolumn{4}{c}{LPIPS $\downarrow$} \\
            \cmidrule{2-19}
            & \multirow{2}{*}{Na\"ive}
            & \multirow{2}{*}{BARF} 
            & \multirow{2}{*}{Ours}
            & \multirow{2}{*}{Na\"ive} 
            & \multirow{2}{*}{BARF} 
            & \multirow{2}{*}{Ours}
            & \multirow{2}{*}{Na\"ive} 
            & \multirow{2}{*}{BARF} 
            & \multirow{2}{*}{Ours} 
            & ref.
            & \multirow{2}{*}{Na\"ive} 
            & \multirow{2}{*}{BARF} 
            & \multirow{2}{*}{Ours} 
            & ref.
            & \multirow{2}{*}{Na\"ive} 
            & \multirow{2}{*}{BARF} 
            & \multirow{2}{*}{Ours} 
            & ref. \vspace{-2.5pt} \\
            & & &
            & & &
            & & & & NeRF
            & & & & NeRF
            & & & & NeRF \\
            \midrule
            Toys & 14.22 & 179.73 & \bf{0.42} & 6.14 & 24.84 & \bf{0.33} & 15.55 & 11.29 & \bf{29.58} & 32.90 & 0.57 & 0.49 & \bf{0.94} & 0.96 & 0.50 & 0.77 & \bf{0.06} & 0.04 \\
            Foods & 5.30  & 10.99 & \bf{0.31} & 7.76  & 10.15 & \bf{0.62} & 19.11 & 18.02 & \bf{31.83} & 24.58 & 0.71 & 0.68 & \bf{0.95} & 0.89 & 0.23 & 0.26 & \bf{0.05} & 0.13 \\
            \bottomrule 
        \end{tabular}
    }
    \vspace{-6pt}
    \caption{
        Quantitative results of bundle-adjusting neural radiance fields on real-world scenes captured using an iPhone under large displacements (\textit{toys}) or sparse views (\textit{foods}).
        L2G-NeRF outperforms baselines and even achieves better performance than reference NeRF that trained under \SFM poses in the Foods scene, which is hard for \SFM to find accurate camera poses.
        Translation errors are scaled by $100$.
    }
    \vspace{-10pt}
    \label{table:nerf-iphone}
\end{table*}

\section{Ablation Studies}

We propose a local-to-global registration method that combines the benefits of parametric and non-parametric methods. The key idea is to apply a pixel-wise alignment that optimizes photometric reconstruction errors $ \sum_{i=1}^{M} \sum^{N}_{j=1} \big\|\R(\T^j_i \x^j; \bTheta) - \I_i(\x^j)\big\|_2^2$, followed by a frame-wise alignment $\sum_{i=1}^{M} \sum^{N}_{j=1} \lambda \big\|\T^j_i \x^j-\T^\ast_i \x^j\big\|_2^2$ to globally constrain the geometric transformations. We evaluate our proposed L2G-NeRF against an ablation (w/o $\mathcal{L}_{global}$), which builds upon our full model by eliminating the global alignment objective, \ie, $\lambda=0$. The ablation is equivalent to a local registration method, while BARF is the chosen representative global registration method.

\begin{figure*}[t]
\centering
  \includegraphics[width=\linewidth]{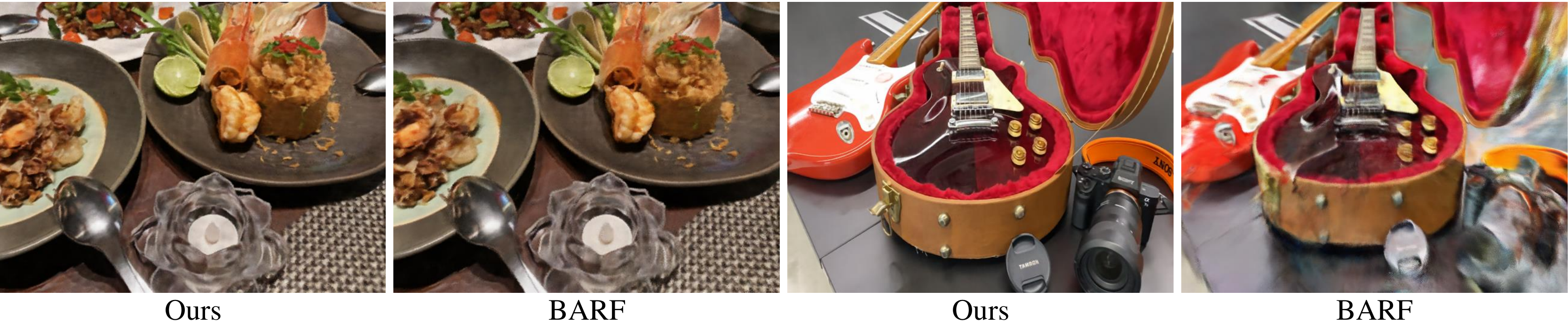}
    \vspace{-12pt}
  \caption{Results of NeRF on reflective scenes.(Shiny dataset)}
\label{fig:reflective}
\end{figure*}

\subsection{Ablation on NeRF (3D): Synthetic Objects}
We first investigate the ablation study of learning NeRF from imperfect camera poses. 
We experiment with 8 synthetic object-centric scenes~\cite{mildenhall2020nerf}. 
The results in Fig.~\ref{fig:ab_synthetic} and Table~\ref{table:nerf-ablation} show that L2G-NeRF achieves better performance than the ablation w/o $\mathcal{L}_{global}$.
Fig.~\ref{fig:ab_synthetic_camera} further illustrates that L2G-NeRF can achieve near-perfect registration while the ablation w/o $\mathcal{L}_{global}$ suffers from suboptimal solutions.

\subsection{Ablation on NeRF (3D): Real-World Scenes}

We further explore the ablation study of learning NeRF in real-world scenes with \emph{unknown} camera poses.
We evaluate on the standard LLFF dataset~\cite{mildenhall2019local}.
Quantitative results are summarized in Table \ref{table:nerf-ablation}. The ablation w/o $\mathcal{L}_{global}$ diverges to wrong poses
(visualized in Fig.~\ref{fig:ab_llff_camera}), 
producing ghosting artifacts (shown in Fig.~\ref{fig:ab_llff}). L2G-NeRF outperforms the ablation w/o $\mathcal{L}_{global}$ and achieves high-quality view synthesis that is competitive to the reference NeRF.

\subsection{Ablation on Neural Image Alignment (2D)}

We further concrete analysis on the homography image alignment experiment and visualize the results in Fig.~\ref{fig:ab_cat}. 
Alignment with w/o $\mathcal{L}_{global}$ results in distorted artifacts (cat ears) in the recovered neural image due to ambiguous registration. 
This is the consequence of w/o $\mathcal{L}_{global}$'s attempt to directly optimize the pixel agreement metric, which minimizes photometric errors but does not obey the geometric constraint (global alignments).
As L2G-NeRF discovers precise warps, it optimizes neural image with high fidelity.

\section{Additional Results} 

\subsection{NeRF (3D): Synthetic Objects}
We report additional qualitative results of learning NeRF from noisy camera poses for synthetic objects in Fig.~\ref{fig:result_synthetic}. The baselines still perform poorly, while L2G-NeRF can achieve near-perfect registration (reflected in Fig.~\ref{fig:result_synthetic_camera}) and render images with comparable visual quality against reference NeRF that trained under ground-truth poses.

\subsection{NeRF (3D): Real-World Scenes (LLFF)}
We report additional qualitative results of learning NeRF for the standard LLFF dataset in Fig.~\ref{fig:result_llff}, where camera poses are \emph{unknown}.
L2G-NeRF successfully recovers the 3D scene with higher fidelity than baselines.
Fig.~\ref{fig:result_llff_camera2} shows that the recovered camera poses from L2G-NeRF agree more with those estimated from \SFM methods than BARF.

\subsection{NeRF (3D): Real-World Scenes (Ours)}
We take one step further to experiment with images captured using an iPhone under 
challenging camera pose distribution.
Fig.~\ref{fig:iphone_camera1} and Fig.~\ref{fig:iphone_camera2} indicate the advantage of L2G-NeRF in registering images captured under large displacements and sparse views, while baselines exhibit artifacts (Fig.~\ref{fig:iphone_nerf1} and Fig.~\ref{fig:iphone_nerf2}) due to unreliable registration, which is reflected in Table \ref{table:nerf-iphone}.
Moreover, the difficulty of registering from sparse views prevents \SFM from finding accurate poses, which results in broken stripes on the synthesis of reference NeRF trained under \SFM poses in \textit{foods} scene.
This further demonstrates the effectiveness of removing the requirement of pre-computed \SFM poses.
Fig.~\ref{fig:iphone_camera1} and Fig.~\ref{fig:iphone_camera2} show the largest displacements (hierarchical but adjacent camera poses) and the sparsest camera setting (9 views) of L2G-NeRF to register images in these scenes successfully, than which we can not handle a more challenging camera pose distribution. 

\subsection{NeRF (3D): Real-World Scenes (Shiny)}
To analyze the influence of reflective surfaces, 
We present an example in \cref{fig:reflective} that reconstructs scenes ~\cite{Wizadwongsa2021NeX} with reflections from identity pose initialization (L2G-NeRF converges, BARF fails in the \textit{guitars} scene).
Interestingly, the global alignment loss increases by 4 to 10 times \wrt other datasets. 
This may be caused by inaccurate local registration in specular regions, 
and our convergence benefits from the global registration constraint.
Specific methods (e.g., Ref-NeRF~\cite{verbin2022ref}) could be employed to handle reflective surfaces better.

{\small
\bibliographystyle{ieee_fullname}
\bibliography{egbib}
}

\end{document}